\documentclass[screen,prologue,dvipsnames,table,xcdraw,sigconf]{acmart}
\AtBeginDocument{%
  \providecommand\BibTeX{{%
    \normalfont B\kern-0.5em{\scshape i\kern-0.25em b}\kern-0.8em\TeX}}}

\copyrightyear{2023}
\acmYear{2023}
\setcopyright{acmlicensed}\acmConference[AIES '23]{AAAI/ACM Conference on AI, Ethics, and Society}{August 8--10, 2023}{Montréal, QC, Canada}
\acmBooktitle{AAAI/ACM Conference on AI, Ethics, and Society (AIES '23), August 8--10, 2023, Montréal, QC, Canada}
\acmPrice{15.00}
\acmDOI{10.1145/3600211.3604707}
\acmISBN{979-8-4007-0231-0/23/08}

\acmConference[AIES '23]{AIES '23: 6th AAAI/ACM Conference on AI, Ethics, and Society}{August 8--10, 2023}{Montréal, QC, Canada}

\usepackage[utf8]{inputenc} % allow utf-8 input
\usepackage[T1]{fontenc}    % use 8-bit T1 fonts
\makeatletter
\renewcommand{\sectionautorefname}{\S\,\@gobble}
\renewcommand{\subsectionautorefname}{\S\,\@gobble}
\renewcommand{\subsubsectionautorefname}{\S\,\@gobble}
\makeatother      % hyperlinks
\usepackage{url}            % simple URL typesetting
\usepackage{booktabs}       % professional-quality tables
\usepackage{hyperref}
\usepackage{multirow}
\usepackage{amsfonts}       % blackboard math symbols
\usepackage{nicefrac}       % compact symbols for 1/2, etc.
\usepackage{microtype} % microtypography
\usepackage{colortbl}         % colors
\usepackage{amsthm}
\usepackage{amsmath}
\usepackage{float}
\usepackage[font=scriptsize]{subcaption}
\usepackage{graphicx}
\usepackage{tablefootnote}
\usepackage{xspace}
\usepackage{bm}
\usepackage{pifont}
\usepackage{array}
\usepackage{breakcites}
\usepackage{stfloats}

\usepackage[super]{nth}
\definecolor{ForestGreen}{RGB}{34,139,34}

\newcommand{\eg}{e.g.,\ }

\newcommand{\ie}{i.e.,\ }

\begin{document}

%%
%% The "title" command has an optional parameter,
%% allowing the author to define a "short title" to be used in page headers.
\title{When Fair Classification Meets Noisy Protected Attributes}

%%
%% The "author" command and its associated commands are used to define
%% the authors and their affiliations.
%% Of note is the shared affiliation of the first two authors, and the
%% "authornote" and "authornotemark" commands
%% used to denote shared contribution to the research.

\author{Avijit Ghosh}
\affiliation{%
  \institution{Northeastern University}
  \city{Boston}
  \country{USA}}
\email{ghosh.a@northeastern.edu}

\author{Pablo Kvitca}
\affiliation{%
  \institution{Northeastern University}
  \city{Boston}
  \country{USA}}
\email{kvitca.p@northeastern.edu}

\author{Christo Wilson}
\affiliation{%
  \institution{Northeastern University}
  \city{Boston}
  \country{USA}}
\email{cbw@ccs.neu.edu}

%%
%% By default, the full list of authors will be used in the page
%% headers. Often, this list is too long, and will overlap
%% other information printed in the page headers. This command allows
%% the author to define a more concise list
%% of authors' names for this purpose.
\renewcommand{\shortauthors}{Ghosh, et al.}

%%
%% The abstract is a short summary of the work to be presented in the
%% article.
\begin{abstract}
  The operationalization of algorithmic fairness comes with several practical challenges, not the least of which is the availability or reliability of protected attributes in datasets. In real-world contexts, practical and legal impediments may prevent the collection and use of demographic data, making it difficult to ensure algorithmic fairness. While initial fairness algorithms did not consider these limitations, recent proposals aim to achieve algorithmic fairness in classification by incorporating noisiness in protected attributes or not using protected attributes at all. 
  
  To the best of our knowledge, this is the first head-to-head study of fair classification algorithms to compare attribute-reliant, noise-tolerant and attribute-unaware algorithms along the dual axes of predictivity and fairness. We evaluated these algorithms via case studies on four real-world datasets and synthetic perturbations. Our study reveals that attribute-unaware and noise-tolerant fair classifiers can potentially achieve similar level of performance as attribute-reliant algorithms, even when protected attributes are noisy. However, implementing them in practice requires careful nuance. Our study provides insights into the practical implications of using fair classification algorithms in scenarios where protected attributes are noisy or partially available.
\end{abstract}

%%
%% The code below is generated by the tool at http://dl.acm.org/ccs.cfm.
%% Please copy and paste the code instead of the example below.
%%
\begin{CCSXML}
<ccs2012>
    <concept>
       <concept_id>10010147.10010257.10010321</concept_id>
       <concept_desc>Computing methodologies~Machine learning algorithms</concept_desc>
       <concept_significance>500</concept_significance>
    </concept>
   <concept>
       <concept_id>10003456.10010927</concept_id>
       <concept_desc>Social and professional topics~User characteristics</concept_desc>
       <concept_significance>500</concept_significance>
    </concept>
   <concept>
       <concept_id>10002944.10011122.10002945</concept_id>
       <concept_desc>General and reference~Surveys and overviews</concept_desc>
       <concept_significance>500</concept_significance>
    </concept>
 </ccs2012>
\end{CCSXML}

\ccsdesc[500]{Social and professional topics~User characteristics}
\ccsdesc[500]{General and reference~Surveys and overviews}
\ccsdesc[500]{Computing methodologies~Machine learning algorithms}

%%
%% Keywords. The author(s) should pick words that accurately describe
%% the work being presented. Separate the keywords with commas.
\keywords{fairness; classification; protected attributes; evaluation}

%% A "teaser" image appears between the author and affiliation
%% information and the body of the document, and typically spans the
%% page.

% \received{20 February 2007}
% \received[revised]{12 March 2009}
% \received[accepted]{5 June 2009}

%%
%% This command processes the author and affiliation and title
%% information and builds the first part of the formatted document.
\maketitle

\section{Introduction}
\label{sec:intro}

In October 2022, the White House released the Blueprint for an AI Bill of Rights~\cite{aiblillofrights}. This document, like other statements of AI principles~\cite{fjeld2020principled,unescoethicalai,oecdethicalai,ibmethics,microsoftethicalai}, calls for protections against unfair discrimination (colloquially, \textit{fairness}) to be deeply integrated into all AI systems. Researchers and journalists have led the way in this area, both in terms of identifying unfairness in real world systems~\cite{angwin2016machine,buolamwini2018gender,dastin2018amazon,makhortykh2021detecting}, and in the development of machine learning (ML) classifiers that jointly optimize for predictive performance and fairness~\cite{dwork2012fairness,kamishima2012fairness,goel2018non,krasanakis2018adaptive} (for a variety of different definitions of fairness~\cite{hardt2016equality,zafar2017fairness,aghaei2019learning,ustun2019fairness}).

Despite the widespread acknowledgment that fairness is a key component of trustworthy AI, formidable challenges remain to the adoption of fair classifiers in real world scenarios---chief among them being questions about demographic data itself. Many \textit{classical fair classifiers} assume that protected attributes are available at training time and/or testing time~\cite{dwork2012fairness} and that this data is accurate. However, demographic data may be noisy for a variety of reasons, including imprecision in human-generated labels~\cite{davani2022dealing}, reliance on imperfect demographic-inference algorithms to generate protected attributes~\cite{ghosh2021fair}, or the presence of an adversary that is intentionally poisoning demographic data~\cite{DBLP:conf/fat/GhoshJW22}. To attempt to deal with these issues, researchers have proposed \textit{noise-tolerant fair classifiers} that aim to achieve distributional fairness by incorporating the error rate of demographic attributes in the fair classifier optimization process itself~\cite{wang2020robust,mozannar2020fair,celis2021fair}.

In other instances demographic data may not be available at all, which violates the assumptions of both classical and noise-tolerant fair classifiers. This may occur when demographic data is unobtainable (\eg laws or social norms impede collection~\cite{bogen2020awareness,andrus2021we}), prohibitively expensive to generate (\eg when large datasets are scraped from the web~\cite{deng2009imagenet,lin2014microsoft,karkkainen2019fairface}), or when laws disallow the use of protected attributes to train classifiers (\eg direct discrimination~\cite{wilson2021building}). For cases such as these, researchers have proposed \textit{demographic-unaware fair classifiers} that use the latent representations in the feature space of the training data to reduce gaps in classification errors between protected groups, either via assigning higher weights to groups of training examples that are misclassified~\cite{hashimoto2018fairness}, or by training an auxiliary adversarial model to computationally identify regions of misclassification ~\cite{lahoti2020fairness}.

Motivated by this explosion of fundamentally different fair classifiers, we present an empirical, head-to-head evaluation of the performance of 14 classifiers in this study, spread across four classes: two \textit{unconstrained classifiers}, seven classical fair classifiers, three noise-tolerant fair classifiers, and two demographic-unaware classifiers. Drawing on the methodological approach used by \citet{friedler2019comparative} in their comparative study of classical fair classifiers, we evaluate the accuracy, stability, and fairness guarantees (defined as the equal odds difference) of these 14 classifiers across four datasets as we vary noise in the protected attribute (sex). To help explain the performance differences that we observe, we calculate and compare the feature importance vectors for our various trained classifiers. This methodological approach enables us to compare the performance of these 14 algorithms under controlled, naturalistic circumstances in an apples-to-apples manner.

Based on our head-to-head evaluation we make the following key observations:
\begin{itemize}
    \item Two classical fair classifiers, one noise-tolerant fair classifier, and one demographic-unaware fair classifier performed consistently well across all metrics on our experiments.
    \item The best classifier for each case study showed some variability, confirming that the choice of dataset is an important factor when selecting a model.
    \item One demographic-unaware fair classifier was able to achieve equal odds for males and females under a variety of ecological conditions, confirming that demographics are not always necessary at training or testing time to achieve fairness.
\end{itemize}
We release our source code and data\footnote{The code and data for replicating this paper can be found at \url{https://github.com/evijit/Awareness_vs_Unawareness}} so that others can replicate and expand upon our results.

We argue that large-scale, head-to-head evaluations such as the one we conduct in this study are critical for researchers and ML practitioners. Our results act as a checkpoint, informing the community about the relative performance characteristics of classifiers within and between classes. For researchers, this can highlight gaps where novel algorithms are still needed (\eg noise-tolerant and demographic-unaware classifiers, based on our findings) and provide a framework for rigorously evaluating them. For practitioners, our results highlight the importance of thoroughly evaluating many classifiers from many classes before adopting one in practice, and we provide a roadmap for choosing the best classifiers for a given real-world scenario, depending on the availability and quality of demographic data.

Our study proceeds as follows: in \autoref{sec:related} we present a brief overview of the history of fair models and head-to-head performance evaluation. Next, in \autoref{sec:algorithms}, we introduce the 14 classifiers and the metrics we use to evaluate them for predictive performance and fairness. In \autoref{sec:methodology} we present our experimental approach, including the datasets we use for our four case studies. In \autoref{sec:results} we present the results of our experiments and we discuss our findings in \autoref{sec:conclusion}.

\section{Related Work}
\label{sec:related}

We discuss different classes of fair classifiers, their known shortcomings, and how they have been evaluated in the past.

\subsection{Fair Classifiers}
\label{sec:fairclassifiers}

\citet{dwork2012fairness} were one of the first to operationalize the idea of fairness in machine learning classifiers, through their key observation that awareness of demographics is crucial for building models that rectify unfair discrimination and historical inequity. Their work takes the idea of awareness literally, by incorporating protected attributes directly into the model and jointly optimizing for accuracy and fairness. Many subsequent works have built on this foundation by developing versions of classical ML classifiers that incorporate fairness constraints (\eg decision trees, random forests, SVMs, boosting, etc.~\cite{mehrabi2019survey}).

Collectively, we refer to this class of algorithms as classical fair classifiers. They are now widely available to practitioners~\cite{linkedin_2021,ibmaif360,bird2020fairlearn} and have been adopted into real-world systems~\cite{ey_2020}. 

While classical fair classifiers are an important advance over their unconstrained predecessors, they rely on a strong assumption that data about protected attributes is accurate. Unfortunately, this may not be true in practice. For example, in contexts like finance and employment candidate screening, demographic data may not be available due to legal constraints or social norms~\cite{wilson2021building,bogen2020awareness}, yet the need to fairly classify people remains paramount. To bridge this gap, practitioners may infer peoples' protected attributes using human labelers~\cite{basu-2020-lighthouse} or algorithms that take names, locations, photos, etc. as input~\cite{proxy_methodology}. However, work by \citet{ghosh2021fair} demonstrates that these inference approaches produce noisy demographic data, and that this noise obviates the fairness guarantees provided by fair models.

With these limitations in mind, researchers have begun developing what we refer to as noise-tolerant fair classifiers that, as the name suggests, jointly optimize for accuracy and fairness in the presence of uncertainty in the protected attribute data. Approaches include robust optimization that adjusts for the presence of noise in the fairness constraint \cite{wang2020robust}, adjusting the ``fairness tolerance'' value for binary protected groups \cite{lamy2019noise}, using noisy attributes to post-process the outputs for fairness instead of the true attributes under certain conditional independence assumptions \cite{awasthi2019effectiveness}, estimating de-noised constraints that allow for near optimal fairness \cite{celis2021fair}, or a combination of approaches \cite{mozannar2020fair}.

Noise-tolerant fair classifiers, like classical fair classifiers, still rely on the assumption that protected attributes are available at training time. As we discuss in \autoref{sec:intro}, however, there are many real-world contexts when this assumption may be violated. The strongest such impediment is legal, \ie any inclusion of protected attributes in the classifier would be considered illegal direct discrimination.

A different approach for achieving fairness through awareness that is amenable to these strong constraints is embodied by what we refer to as demographic-unaware fair classifiers. These algorithms do not take protected attributes as input, but they attempt to achieve demographic fairness anyway by relying on the latent representations of the training data~\cite{hashimoto2018fairness,lahoti2020fairness}. Thus, this approach to classification still incorporates a general awareness of unfair discrimination and historical inequity without being directly aware of demographics.

While demographic-unaware fair classifiers are an attractive solution in contexts where protected attributes are unavailable, practical questions about the efficacy of these algorithms remain. First, because these techniques are unsupervised, it is unclear what groups are identified for fairness optimization. Under what circumstances are demographic-unaware fair classifiers able to achieve fairness for social groups that have been historically marginalized or are legally protected? Conversely, are the groups constructed by demographic-unaware fair classifiers arbitrary and thus divorced from salient real-world sociohistorical context? Second, assuming that demographic-unaware fair classifiers do identify and act on meaningful groups of individuals, how does their performance (in terms of predictions and fairness) compare to classical and noise-tolerant fair classifiers? In this study, our goal is to begin answering these questions about relative performance across all four classes of fair classifiers.

\subsection{Head-to-Head Evaluation}

It is standard practice for ML researchers to compare the performance of their novel algorithms against competitors. However, these comparisons are rarely comprehensive, \ie they focus on comparisons with a narrow set of comparable algorithms to demonstrate advances over the state-of-the-art. While these evaluations are crucial for assessing the benefits of new algorithms, they do not paint a complete picture of performance across a variety of different algorithms, spanning both time and fundamental approaches.

Benchmark studies address this gap by focusing on the evaluation of a large set of models under expansive and carefully controlled conditions~\cite{friedler2019comparative,hort-2021-fairea}. These studies provide important context for the ML field, \eg by identifying models that do not work well in practice, models that have equivalent performance characteristics under a wide range of circumstances, and areas where new models may be needed. To the best of our knowledge, existing benchmark studies focus solely on classical fair classifiers, which motivates us to update their results. Thus, in this study we adopt the methodological approach for evaluation developed by \citet{friedler2019comparative} and build upon their work by evaluating four different classes of classifiers (both fairness constrained and unconstrained).

\section{Algorithms and Metrics}
\label{sec:algorithms}

In this section, we introduce the 14 classifiers that we evaluated in this study and the metrics we used to evaluate them.

\subsection{Classifiers}

We group the classifiers that we evaluated in this study into four classes: (1) unconstrained classifiers that solely optimize for accuracy; (2) classical fair classifiers that require access to protected attributes at training (and sometimes testing) time, and assume that this data are accurate; (3) noise-tolerant fair classifiers that also require access to protected attributes but account for uncertainty in the data; and (4) demographic-unaware fair classifiers that jointly optimize for accuracy and fairness but without access to any protected attribute data. The set of classifiers we have selected is not exhaustive. Instead, we aim to include representative classifiers from the various types of approaches that exist within each class. We discuss the classifiers from each class that we selected for our study below, with further details on related approaches in each subsection. 

\subsubsection{Unconstrained Classifiers}

We chose two classifiers that do not have any fairness constraints, \ie they only aim to maximize predictive accuracy.

\begin{itemize}
    \item \textbf{Logistic Regression (LR)} is the simplest classifier we evaluate. While LR is demographic-aware because it takes all features (including protected attributes) as model inputs at both train and test time, it is not designed to achieve any fairness criteria.
    \item \textbf{Random Forest (RF)} is an ensemble method for classification built out of decision trees. Like LR, we train RF classifiers on all input features including protected attributes.
\end{itemize}

\subsubsection{Classical Fair Classifiers}

We chose seven classifiers from the literature that take protected attributes as input and attempt to achieve demographic fairness. These classifiers vary with respect to how they implement fairness, \ie by pre-processing data, in-process during model training, or by post-processing the trained model. In particular, there exist many techniques for fairness optimization in this class, such as: reweighting of samples via group sizes \cite{calmon2017optimized,kamiran2012data,feldman2015certifying} or via mutual independence of protected and unprotected features in the latent representations \cite{zhang2018mitigating,zemel2013learning}, adding fairness constraints during the learning process~\cite{kamishima2012fairness,zafar2017fairness, agarwal2018reductions, agarwal2019fair}, or by changing the output labels to match some fairness criterion \cite{pleiss2017fairness,kamiran2012decision}. The seven classifiers we choose below are representative of these different approaches.

\begin{itemize}
    \item \textbf{Sample Reweighting (SREW)} is a pre-processing technique that takes each (group, label) combination in the training data and assigns rebalanced weights to them. The goal of this procedure is to remove imbalances in the training data, with the ultimate aim of ensuring fairness before the classifier is trained~\cite{kamiran2012data}. 
    \item \textbf{Learned Fair Representation (LFR)} is a pre-processing technique that converts the input features into a latent encoding that is designed to represent the training data well while simultaneously hiding protected attribute information from the classifier~\cite{zemel2013learning}. 
    \item \textbf{Adversarial Debiasing (ADDEB)} is an in-process technique that trains a classifier to maximize accuracy while simultaneously reducing an adversarial network's ability to determine the protected attributes from the predictions~\cite{zhang2018mitigating}.
    \item \textbf{Exponentiated Gradient Reduction (EGR)} is an in-process technique that reduces fair classification to a set of cost-sensitive classification problems, essentially treating the main classifier itself as a black box and forcing the predictions to be the most accurate under a given fairness constraint~\cite{agarwal2018reductions}. In this case, the constraint is solved as a saddle point problem using the exponentiated gradient algorithm. 
    \item \textbf{Grid Search Reduction (GSR)} uses the same set of cost-sensitive classification problems approach as EGR, except in this case the constraints are solved using the grid search algorithm~\cite{agarwal2018reductions,agarwal2019fair}.
    \item \textbf{Calibrated Equalized Odds (CALEQ)} is a post-processing technique that optimizes the calibrated classifier score output to find the probabilities that it uses to change the output labels, with an equalized odds objective~\cite{pleiss2017fairness}.
    \item \textbf{Reject Option Classifier (ROC)} is a post-processing technique that swaps favorable and unfavorable outcomes for privileged and unprivileged groups around the decision boundaries with the highest uncertainty~\cite{kamiran2012decision}.
\end{itemize}
Note that the CALEQ and ROC algorithms have access to protected attributes at both train and test time, while the other classifiers only have access to protected attributes at training time.

\subsubsection{Noise-tolerant Fair Classifiers}

We chose three classifiers from the literature that take protected attributes as input and attempt to achieve demographic fairness even in the presence of noise. Other than the three classifiers that we chose, we are aware of only one other approach: by \citet{celis2021fair}, who suggests using de-noised constraints to achieve near-optimal fairness.\footnote{\citet{celis2021fair}'s source code only supported Statistical Parity and False Discovery constraints, not EOD, which is why we omitted their classifier from our analysis.}

\begin{itemize}
    \item \textbf{Modified Distributionally Robust Optimization (MDRO)} by \citet{wang2020robust} is an extension of the Distributionally Robust Optimization (DRO) algorithm \cite{hashimoto2018fairness} that adds a maximum total variation distance in the DRO procedure. By assuming a noise model for the protected attributes, it aims to provide tighter bounds for DRO.
    \item \textbf{Soft Group Assignments (SOFT)}, also by \citet{wang2020robust}, is a theoretically robust approach that first performs ``soft'' group assignments and then performs classification, with the idea being that if an algorithm is fair in terms of those robust criteria for noisy groups, then they must also be fair for true protected groups \cite{kallus2022assessing}.
    \item \textbf{Private Learning (PRIV)} is an approach by \citet{mozannar2020fair} that uses differential privacy techniques to learn a fair classifier while having partial access to protected attributes. The approach requires two steps. The first step is to obtain locally private versions of the protected attributes (like \citet{lamy2019noise}). Second, following \citet{awasthi2019effectiveness}, PRIV tries to create a fair classifier based on the private attributes. For this study, we select the privacy level hyperparameter to be a medium value (zero).
 \end{itemize}

\subsubsection{Demographic-unaware Fair Classifiers}

We chose two classifiers from the literature that attempt to achieve fairness without taking protected attributes as input.  

\begin{itemize}
    \item \textbf{Adversarially Reweighted Learning (ARL)} harnesses non-protected attributes and labels by utilizing the computational separability of these training instances to divide them into subgroups, and then uses an adversarial reweighting approach on the subgroups to improve classification fairness~\cite{lahoti2020fairness}.
    \item \textbf{Distributionally Robust Optimization (DRO)} is an algorithm that attempts to minimize the worst case risk of all groups that are close to the empirical distribution~\cite{hashimoto2018fairness}. In the spirit of Rawlsian distributive justice, the algorithm tries to control the risk to minority groups while being oblivious to their identities.
 \end{itemize}

 These two classifiers operate under similar principles: they both try to reduce the gap in errors between protected groups by reducing the classification errors between latent groups in the training set. They do however have one difference: while DRO just increases the weights of the training examples that have higher errors, ARL trains an auxillary adversarial network to identify the regions in the latent input space that lead to higher errors and tries to equalize them, a phenomenon \citet{lahoti2020fairness} call \textit{computational identifiability}.
 
\subsection{Evaluation Metrics}\label{sec:metrics}

To compare the above 14 classifiers head-to-head, we studied their predictive power and their ability to achieve a fairness condition. We also measured the stability of these quantities when noise in the protected attributes was and was not present (described in \autoref{sec:synthetic}).

To assess predictive performance we computed accuracy, defined as:
\begin{equation}
    \scriptstyle
    \text{Accuracy} = \frac{ \scriptstyle \text{number of correct classifications}}{ \scriptstyle \text{test dataset size}}.
\end{equation}
Accuracy is continuous between zero and one with the ideal value being one, which indicates a perfectly predictive classifier.

Many measures of fairness exist in the literature \cite{mehrabi2019survey}. For the purposes of this study, however, we needed to choose a metric that is supported by all the 14 classifiers so that our comparison is apples-to-apples. The classical and noise-tolerant fair classifiers have support for achieving any user-specified fairness constraint, while the demographic-unaware fair classifiers try to minimize the gap in utility between the protected groups. Based on this limitation, and for the sake of brevity, we choose the Average Odds Difference between two demographic groups as our fairness metric, and subsequently choose Equal Odds Difference (EOD) over both groups as our regularization constraint for the classical and noise-tolerant fair classifiers. EOD is defined as:
\begin{equation}
    { \scriptstyle
        \text{EOD} = \frac{\scriptstyle ({\text{FPR}}_{\text{unpriv}} - {\text{FPR}}_{\text{priv}})+({\text{TPR}}_{\text{unpriv}} - {\text{TPR}}_{\text{priv}})}{
        \scriptstyle 2}
    }
\end{equation}
\label{eq:eod}
\noindent where $\text{TPR}$ is the true positive rate and $\text{FPR}$ is the false positive rate. $\text{Priv}$ and $\text{Unpriv}$ denote the privileged and unprivileged groups, respectively. The ideal value of EOD is zero, which indicates that both groups have equal odds of correct and incorrect classification by the trained classifier.

In this study, when we evaluate fairness, we do so for binary sex attributes. We adopted this approach because the datasets we use in our evaluation all include this attribute (see \autoref{sec:methodology}) and four classifiers in our evaluation (\eg CALEQ, ROC, EGR, GSR) only support fairness constraints over two groups. Whenever necessary, we consider males to be the privileged group and females to be the unprivileged group. Note that optimizing for fairness between two groups is the simplest scenario that fair classifiers will encounter in practice---if they perform poorly on this task, then they are unlikely to succeed in more complex scenarios with multiple, possibly intersectional, groups.

\begin{figure*}[t]
    \includegraphics[width=\textwidth,keepaspectratio]{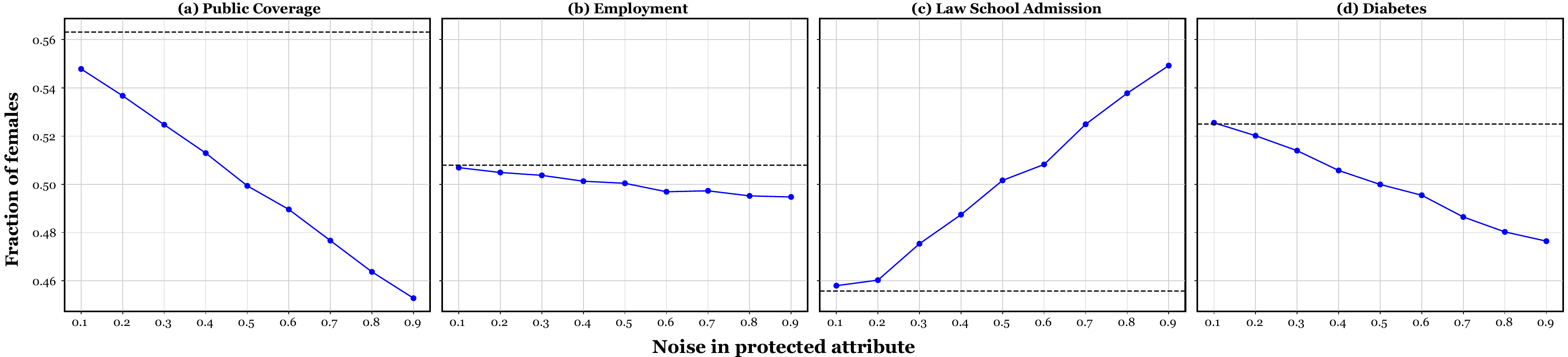}
    \caption{Fraction of females in our datasets after adding synthetic noise. The dashed line indicates the true fraction of females.}
    \label{fig:fractionplot}
\end{figure*}

\section{Methodology}
\label{sec:methodology}

In this section, we describe the approach we used to empirically evaluate the 14 classifiers that we chose for our study. 

\subsection{Case Studies}
\label{sec:cs}

To observe how the classifiers perform on real-world data we chose four different datasets. The classification tasks are described below. Each dataset had binary sex as part of the input features. 

\begin{enumerate}
    \item \textbf{Public Coverage~\cite{ding2021retiring}.} The task is to predict whether an individual (who is low income and not eligible for Medicare) was covered under public health insurance. We used census data from California for the year 2018.
    \item \textbf{Employment~\cite{ding2021retiring}.} The task is to predict whether an individual (between the ages of 16 and 90), is employed. For this task too, we looked at census data from California for the year 2018.
    \item \textbf{Law School Admissions~\cite{wightman1998lsac}.} The task is to predict whether a student was admitted to law school.
    \item \textbf{Diabetes~\cite{strack2014impact}.} The task is to predict whether a diabetes patient was readmitted to the hospital for treatment after 30 days.
\end{enumerate} 

For each of these case studies, we split the dataset into train and test sets in an 80:20 ratio, trained every classifier on the same training set, and then used the trained classifiers to generate predictions on the same testing set. We verified via two-tailed Kolmogorov–Smirnov tests \cite{kolmogorov1933sulla,smirnov1939estimation} and Mann–Whitney $U$ tests \cite{mann1947test} that the test set distribution for every feature was the same as the training set distribution. Finally, we calculated the metrics in \autoref{sec:metrics} on these predictions and compared the results from each classifier head-to-head. We repeated this procedure ten times to assess the stability of accuracy and EOD for each classifier.

\subsection{Synthetic Noise}
\label{sec:synthetic}

While studying the performance of these classifiers on a variety of real-world datasets is important, in order to get a more thorough understanding of the theoretical fairness and predictivity limits of the classifiers we subjected them to robust synthetic stress tests. As discussed in \autoref{sec:fairclassifiers}, in the real world, practitioners may not have access to the protected attribute information of people in their dataset. As a result, practitioners may use inference tools to find proxies for protected attributes, which can lead to unexpected, unfair outcomes \cite{ghosh2021fair}. To characterize what might happen in such a scenario, we perform the following synthetic experiments:
\begin{enumerate}
    \item For each dataset, with a given probability (ranging from 0.1 to 0.9), we randomly flip the protected attribute labels (binary sex in this case) in the dataset. We refer to this probability value as \textit{noise}.
    \item With the synthetically generated dataset from Step 1, we then proceed to split the dataset 80:20, train all 14 algorithms on the same training set, and then calculate predictions on the same test set. The noisy (flipped) labels are passed as inputs to the classifiers at this step.
    \item Next, with the predicted outcomes from Step 2, we calculate accuracy and EOD. Note that we calculate EOD with the \textit{true} protected attributes, \ie we measure the output bias in terms of the original sex labels from the given dataset.
    \item We repeat Steps 1--3 ten times for each value of noise, to ensure statistical fairness and assess the stability of our metrics per classifier.
\end{enumerate}

\autoref{fig:fractionplot} shows the fraction of females in the noised datasets at each level of noise. The fraction of females goes up or down with noise depending on what the true fraction of females in the different datasets were to begin with.

 \begin{figure}[t]
    \includegraphics[width=0.8\columnwidth,keepaspectratio]{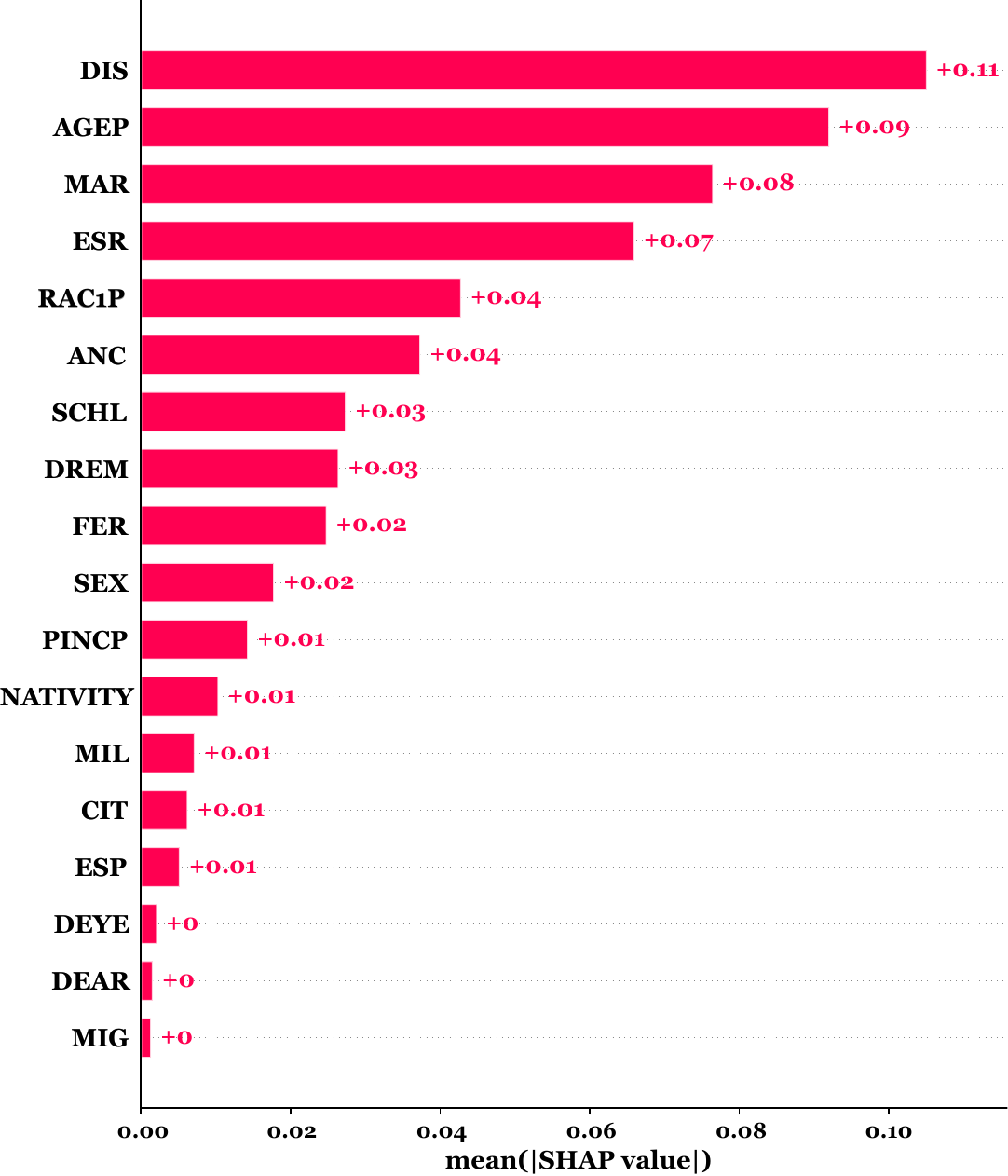}
    \caption{KernelShap feature explanations calculated for the Logistic Regression (LR) classifier when trained on the Public Coverage dataset with no added noise. We used the same approach to calculate feature importances for every classifier-dataset pair at different noise levels.}
    \label{fig:kernelshap}
\end{figure}

\subsection{Calculating Feature Importance}
\label{sec:meth:exp}

To help explain the variations in performance that we observed in our results, we calculated feature importance for each of our trained models. Although there are several black-box model explanation tools in the research literature---such as LIME~\cite{ribeiro2016should}, SHAP~\cite{lundberg2017unified}, and Integrated Gradients~\cite{sundararajan2017axiomatic}---we required an explanation method that was model agnostic. The method that we settled on was KernelShap.\footnote{\url{https://shap-lrjball.readthedocs.io/en/latest/generated/shap.KernelExplainer.html}} According to the documentation, KernelShap uses a special weighted linear regression model to calculate local coefficients, to estimate the Shapley value (a game theoretic concept that estimates the individual contribution of each player towards the final outcome). As opposed to retraining the model with every combination of features as in vanilla SHAP, KernelShap uses the full model and integrates out different features one by one. It also supports any type of model, not just linear models, and was thus a good candidate for our study.

\autoref{fig:kernelshap} shows an example distribution of feature importances calculated for the LR algorithm when trained on the Public Coverage dataset at noise level zero (\ie no noise). In a similar fashion, we used KernelShap to calculate feature importance values for trained classifier outputs at noise levels 0, 0.2, 0.4, 0.6 and 0.8 for all 14 models. 

Research by \citet{kumar2020problems} has shown that different explanation methods often do not agree with each other. We do not claim that the feature importances we calculated using KernelShap are guaranteed to agree with those produced by other tools. Nonetheless, we are specifically interested in the relative importance of the sex feature towards the final outcome as compared to the other input features. Shapley value-based explanations give us a reasonable sense of relative feature importance, as has been empirically shown in previous work \cite{ghosh2022faircanary}. 

\begin{figure*}[t]
    \centering
    \begin{subfigure}[t]{0.267\textwidth}
        \vskip 0pt
        \includegraphics[width=\columnwidth,keepaspectratio]{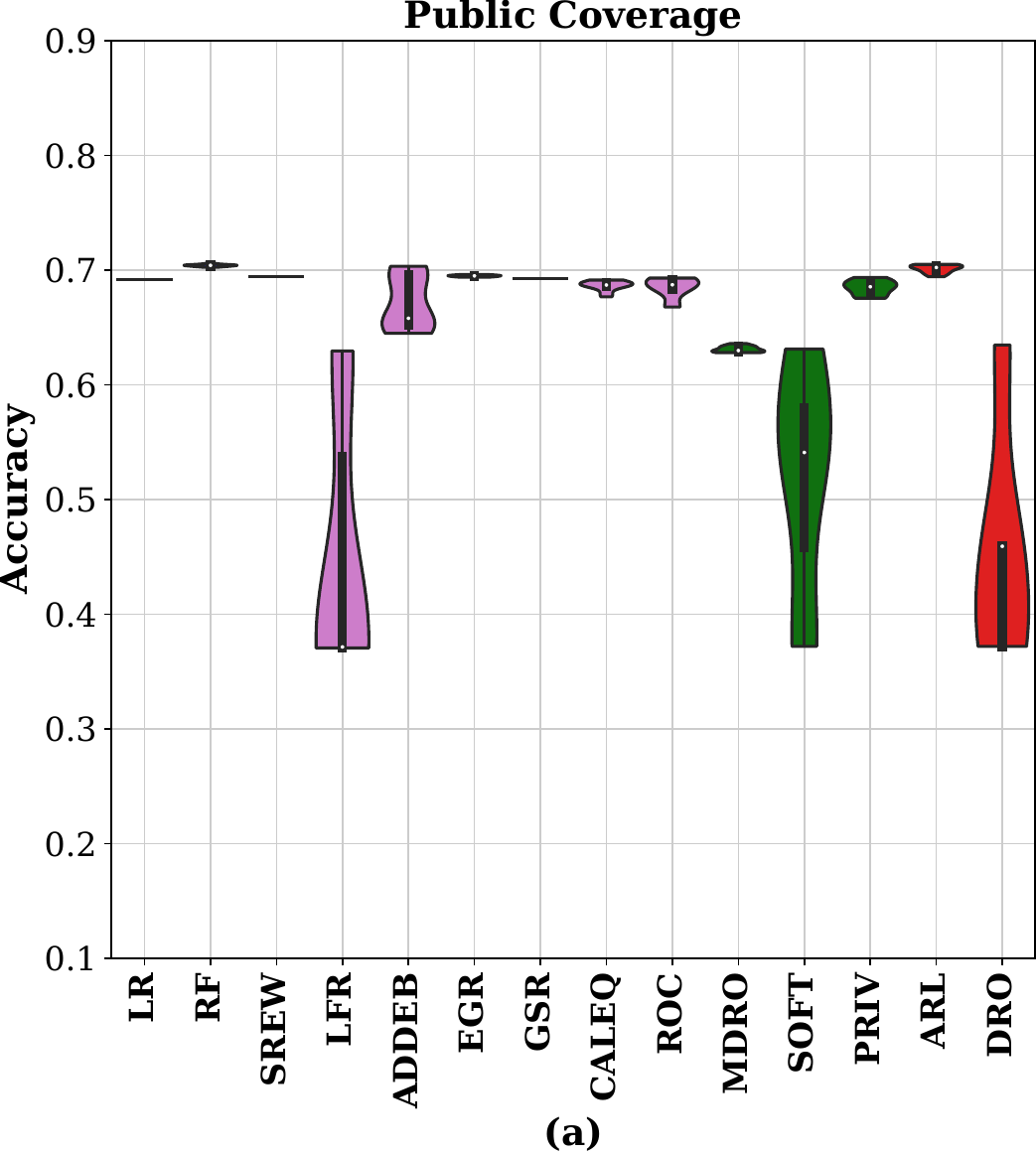}
    \end{subfigure}
    \hfill
    \begin{subfigure}[t]{0.238\textwidth}
        \vskip 0pt
        \includegraphics[width=\columnwidth,keepaspectratio]{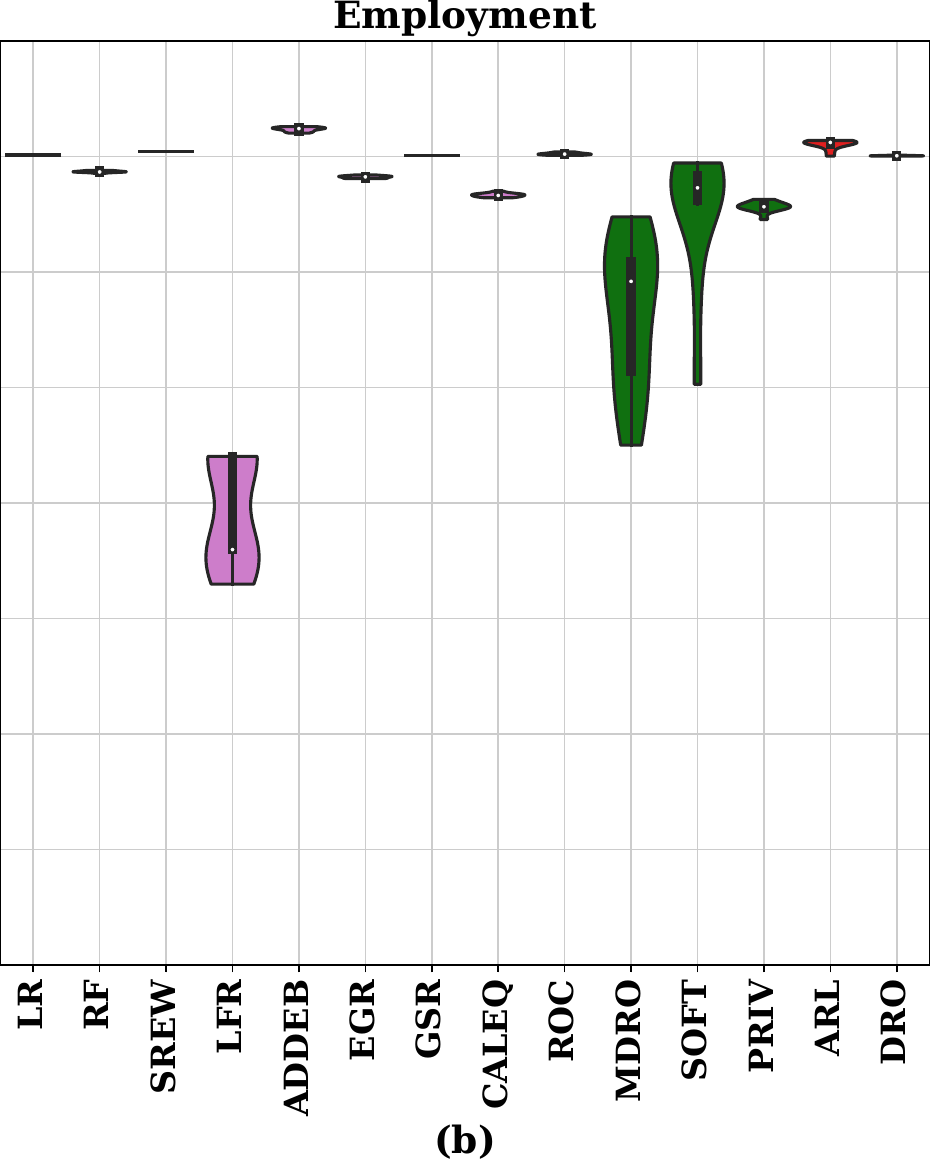}
    \end{subfigure}
    \hfill
    \begin{subfigure}[t]{0.238\textwidth}
        \vskip 0pt
        \includegraphics[width=\columnwidth,keepaspectratio]{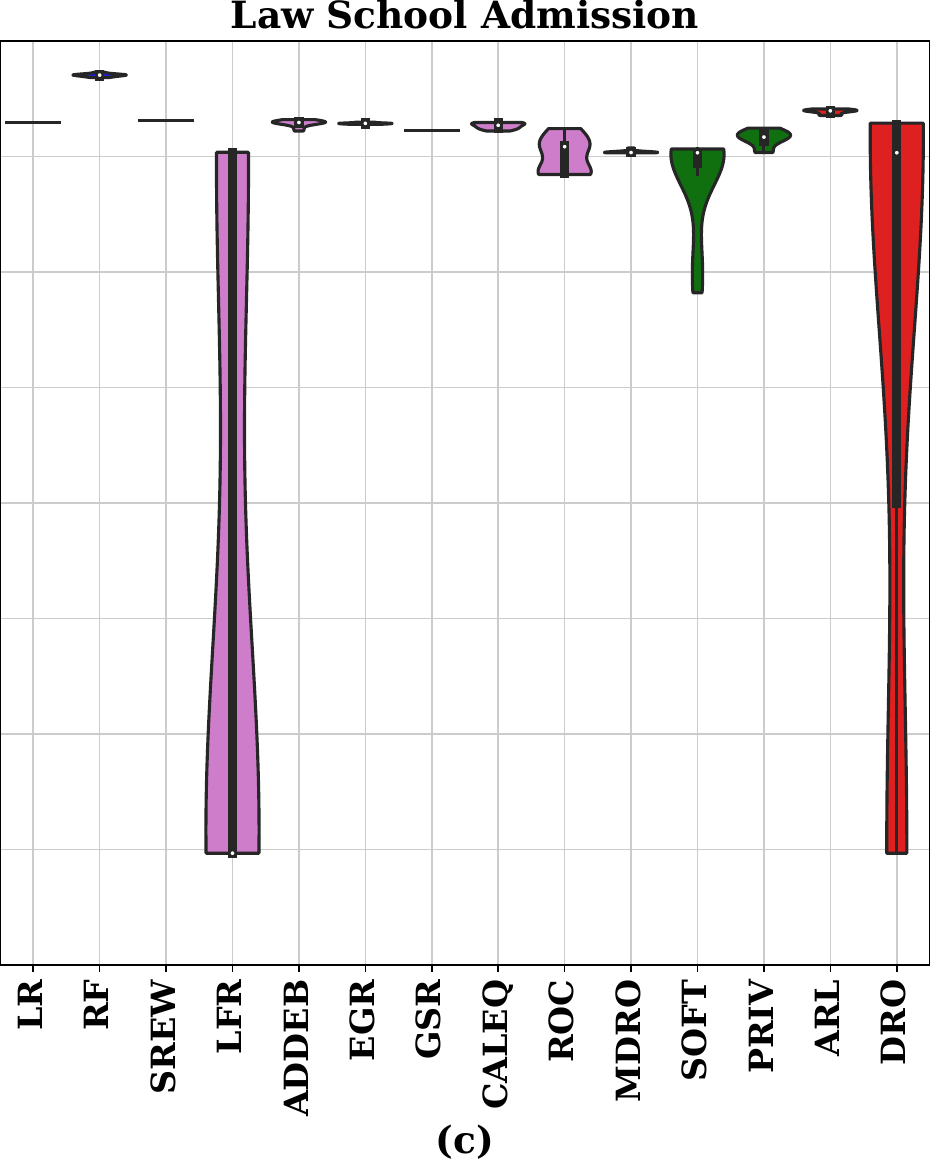}
    \end{subfigure}
    \hfill
    \begin{subfigure}[t]{0.238\textwidth}
        \vskip 0pt
        \includegraphics[width=\columnwidth,keepaspectratio]{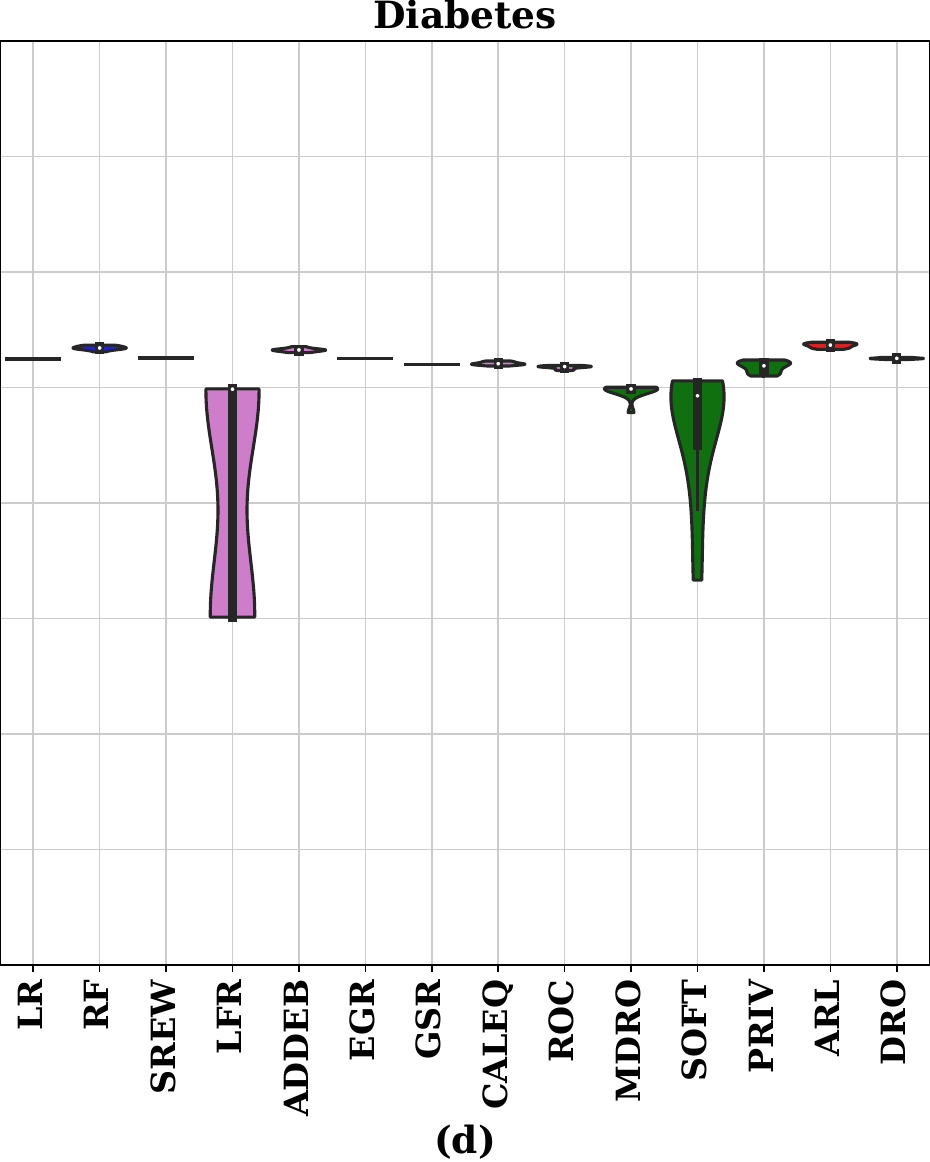}
    \end{subfigure}
    \vskip 2pt
    \vfill
    \begin{subfigure}[t]{0.276\textwidth}
        \vskip -1pt
        \hspace*{-7pt}
        \includegraphics[width=\columnwidth,keepaspectratio]{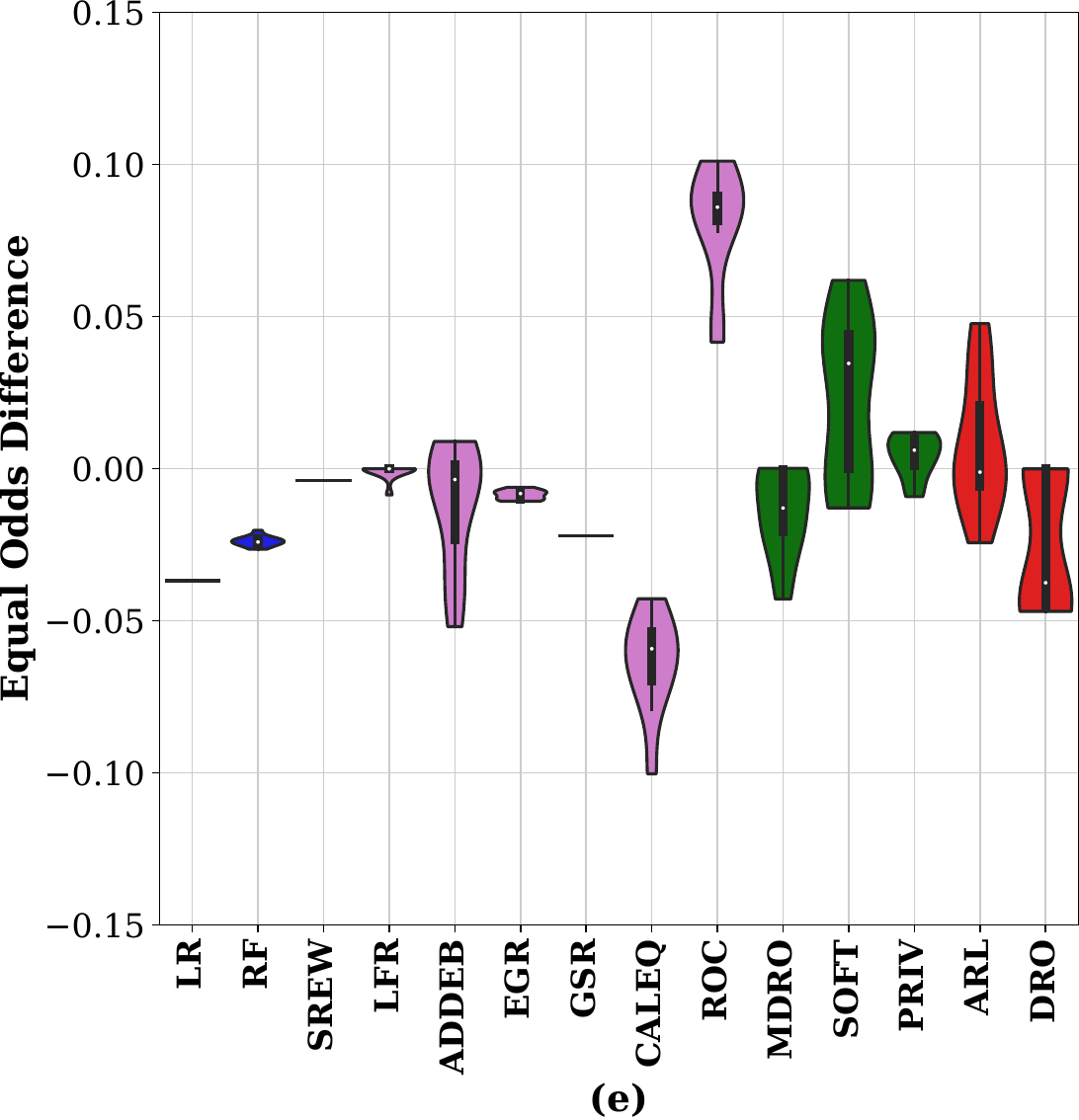}
    \end{subfigure}
    \hfill
    \begin{subfigure}[t]{0.237\textwidth}
        \vskip 0pt
        \hspace*{-6pt}
        \includegraphics[width=\columnwidth,keepaspectratio]{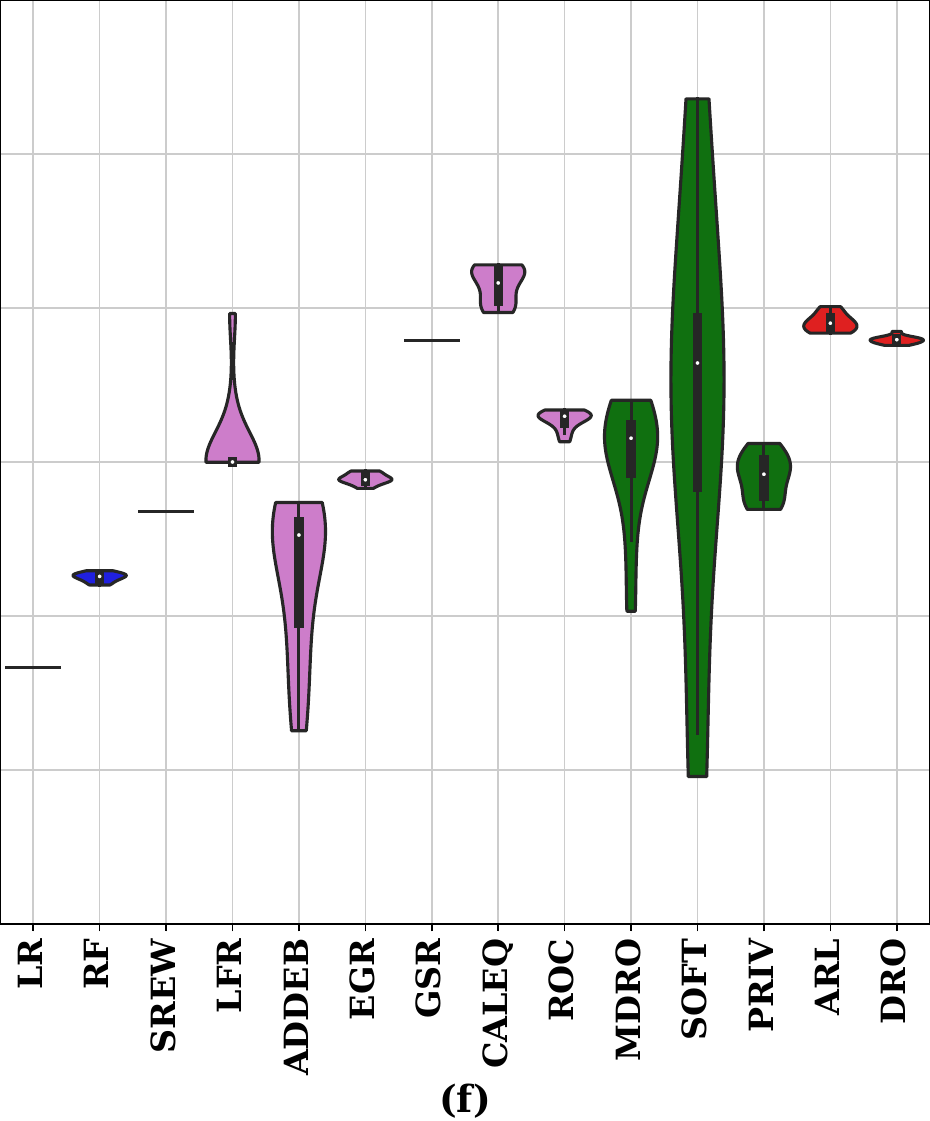}
    \end{subfigure}
    \hfill
    \begin{subfigure}[t]{0.237\textwidth}
        \vskip 0pt
        \hspace*{-4pt}
        \includegraphics[width=\columnwidth,keepaspectratio]{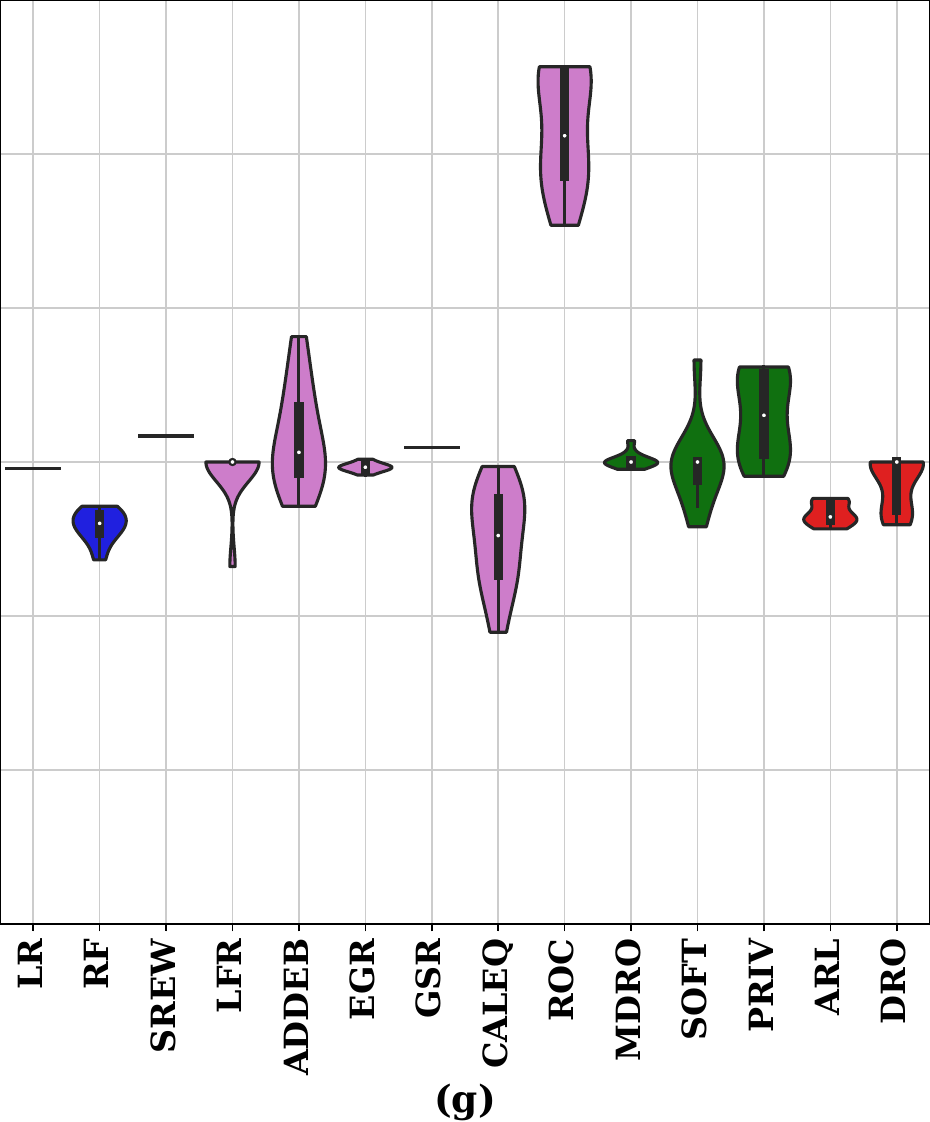}
    \end{subfigure}
    \hfill
    \begin{subfigure}[t]{0.237\textwidth}
        \vskip 0pt
        \hspace*{-3pt}
        \includegraphics[width=\columnwidth,keepaspectratio]{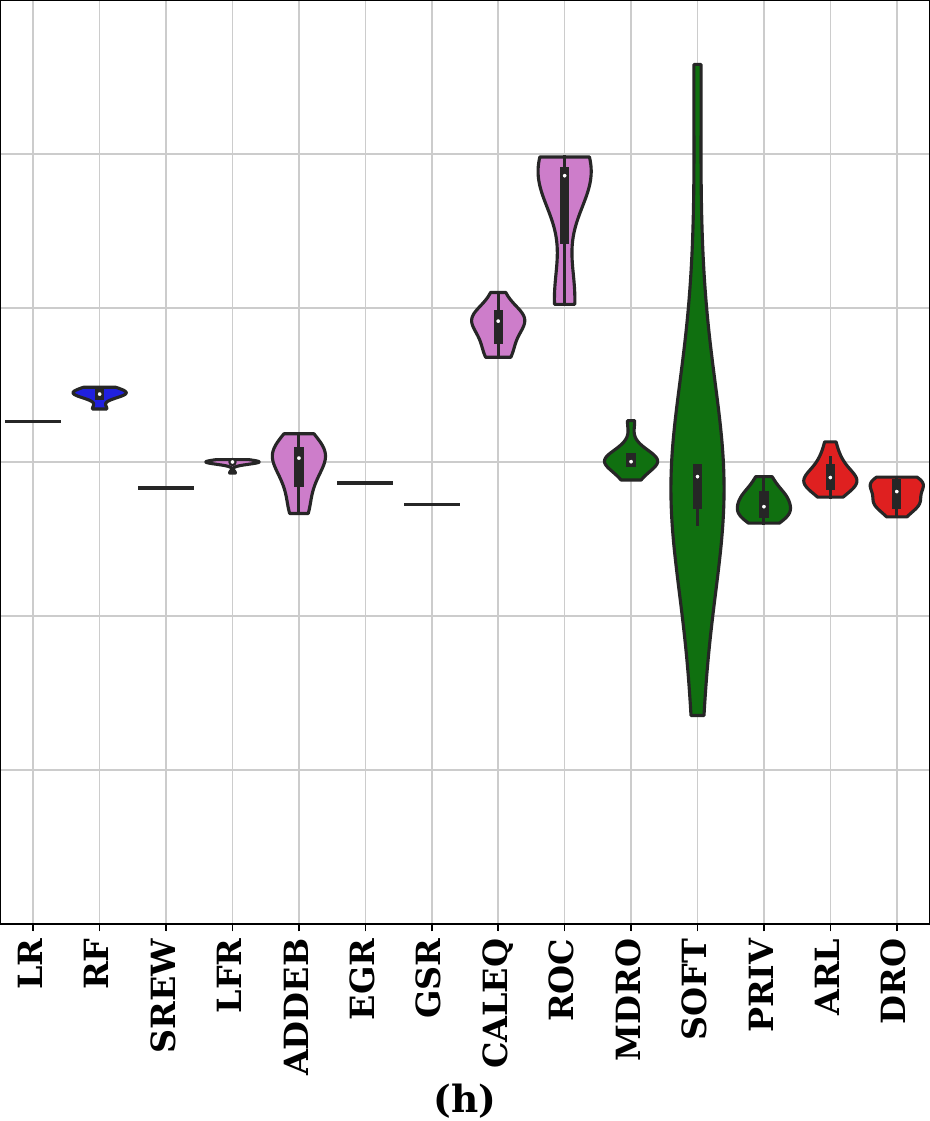}
    \end{subfigure}
    \vfill
    \caption{Accuracy and EOD for our 14 classifiers, calculated over four datasets with ten runs each. No noise was added to the protected attribute in these tests. Violins are color coded by class: blue for unconstrained classifiers, purple for classical fair classifiers, green for noise-tolerant fair classifiers, and red for demographic-unaware fair classifiers. LR, SREW, and GSR are deterministic algorithms and therefore appear as fixed points.}
    \label{fig:baseline_stability}
\end{figure*}

\section{Results}
\label{sec:results}

In this section, we present the results of our experiments. We begin by examining the baseline performance of the 14 classifiers when there is no noise, followed by their performance in the presence of synthetic noise. Finally, we delve into feature importance explanations to help explain the relative performance characteristics of the classifiers.

\subsection{Baseline Characteristics}

\autoref{fig:baseline_stability}(a--d) shows the accuracy and fairness outcomes for all 14 classifiers when there was no noise in the datasets. We executed each classifier ten times without fixing a random seed and present the resulting distributions of metrics using violin plots. We observe that most of the classifiers achieved comparable accuracy to each other on each dataset, and that most classifiers exhibited stable accuracy over the ten executions of the experiments. Learned Fair Representation (LFR), Soft Group Assignment (SOFT), and Distributed Robust Optimization (DRO) were the exceptions: the former two exhibited unstable accuracy on all four datasets, the latter on two datasets. 

As shown in \autoref{fig:baseline_stability}(e--h), EOD was considerably more variable over runs than accuracy. The unconstrained classifiers (LR and RF) were relatively stable and, in some cases, achieved roughly equalized odds (\eg on the Law School and Diabetes datasets). The classical fair classifier group contained the two least fair classifiers in these experiments (CALEQ and ROC), while the other pre-processing and in-processing algorithms performed relatively better. Adversarial Debiasing (ADDEB) was slightly unstable but the distribution centered around zero. Among the noise-tolerant fair classifiers, Soft Group Assignment (SOFT) was unstable on three out of four datasets, while the other two classifiers (MDRO and PRIV) were relatively more stable and more fair. The two demographic-unaware fair classifiers (ARL and DRO) were unstable on the Public Coverage dataset (\autoref{fig:baseline_stability}e) and did not achieve equalized odds on the Employment dataset (\autoref{fig:baseline_stability}f). However, ARL and DRO were stable and fair on the remaining two datasets.

% In a pleasant surprise, the two \textit{Demographic-unaware Fair} classifiers DRO and ARL seem to be performing adequately in terms of fairness. Given that neither of these two algorithms have access to sex in neither the train or test set, we attempt to hypothesize what happened. We first of all observe that DRO has poor accuracy and is jittery in \autoref{fig:lineplots-synthetic}a,c. Since DRO assigns higher weights to groups of training examples that are misclassified, it is susceptible to outliers. This is what we hypothesize happened -- in the presence of large outliers in the two training datasets (Public Coverage and Law School), DRO failed to learn an accurate function and instead predicted all zeroes in the output, which is reflected in its stability plots (\autoref{fig:baseline_stability}a,c and \autoref{fig:stab-noise}a,e). However, when all outcomes are predicted as 0, it leads to a value of perfect fairness, since every group is being (mis)treated equally! ARL escapes this fate, since the internal adversarial network it trains to identify regions of higher errors is resistant to outliers, as detailed in \cite{lahoti2020fairness}. We observe the benefits of this approach in the outcomes, as ARL is able to consistently achieve good accuracy and fairness levels over all datasets.

In summary, we observe that the accuracy and fairness performance of these classifiers was dependent on the dataset that they are trained and tested on, \ie there was no single best classifier. Additionally, we can see that several classifiers are consistently unstable, which explains some of the results that we will present in the next section.

\begin{figure*}[t]
    \centering
    \begin{subfigure}[t]{\textwidth}
        \vskip 0pt
        \includegraphics[width=\columnwidth,keepaspectratio]{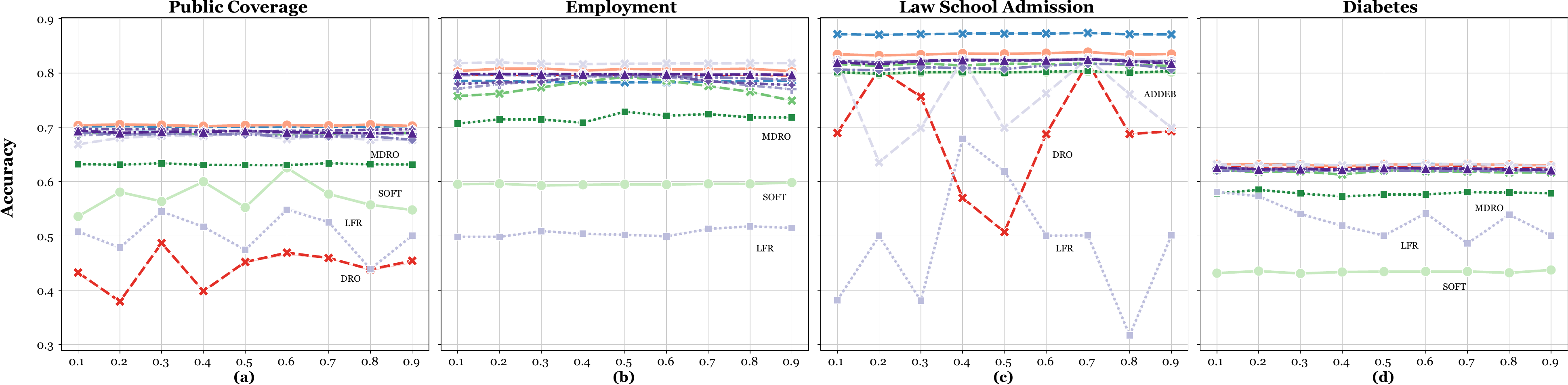}
    \end{subfigure}
    \vfill
    \begin{subfigure}[t]{\textwidth}
        \vskip 0pt
        \includegraphics[width=\columnwidth,keepaspectratio]{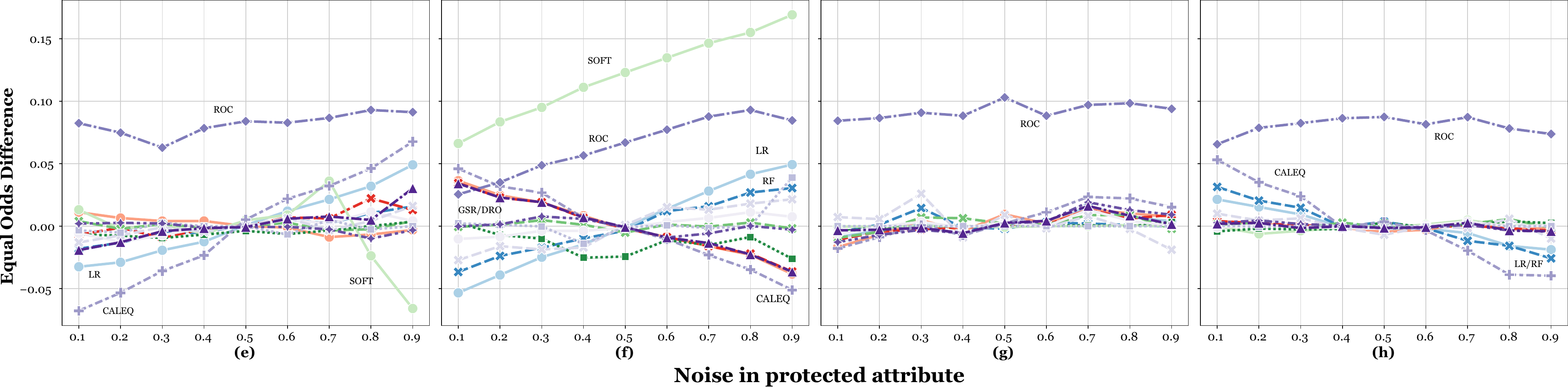}
    \end{subfigure}
    \vfill
    \vspace{5pt}
    \begin{subfigure}[t]{0.45\textwidth}
        \vskip 0pt
        \hspace{8pt}
        \includegraphics[width=\columnwidth,keepaspectratio]{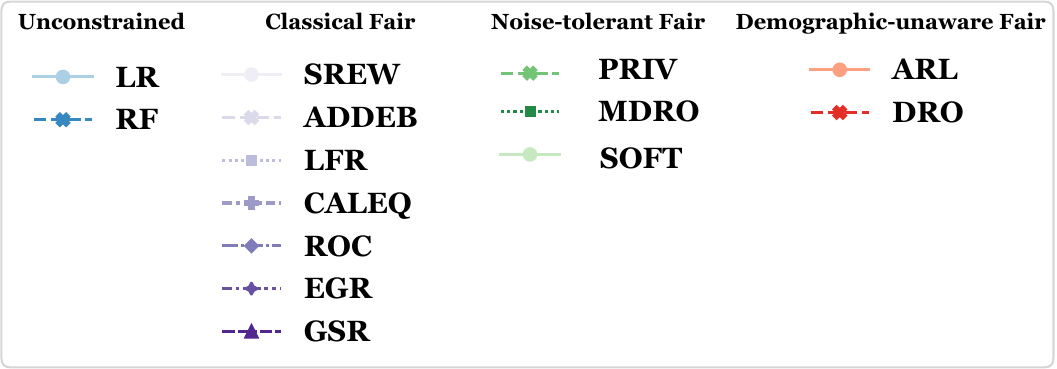}
    \end{subfigure}
    \vfill
    \caption{Accuracy and EOD for our 14 classifiers, calculated over four datasets as we increase noise in the protected attribute (sex). Each point is the average of ten runs for a given classifier, dataset, and noise level. Classifiers are color coded according to the legend. We highlight classifiers whose performance significantly diverges from the consensus with annotated labels.}
    \label{fig:lineplots-synthetic}
\end{figure*}

\subsection{Characteristics Under Noise}

Next, we present the results of experiments where we added noise to the protected attribute of the datasets. We added noise in increments of 0.1 starting from 0.1 and ranging up to 0.9. We added a given amount of noise to each dataset ten times and repeated the experiment, thus we plot the average values of accuracy and EOD for each classifier at each noise level. 

\autoref{fig:lineplots-synthetic}(a--d) shows the accuracy of the 14 classifiers' outputs as we varied noise. We observe that the MDRO, SOFT, and LFR classifiers had poor accuracy across all datasets and noise levels, while the DRO classifier had poor accuracy in two out of the four datasets. These observations mirror those from \autoref{fig:baseline_stability}, \ie these classifiers exhibited poor average accuracy in the noisy experiments because they were unstable in general. The other classifiers tended to be both accurate and stable, irrespective of noise.

As shown in \autoref{fig:lineplots-synthetic}(e--h), the EOD results were much more complex than the accuracy results. ROC generated unfair outputs over all four datasets, at every noise level. Its companion post processing algorithm, CALEQ, exhibited rising EOD with noise for the Public Coverage dataset (\autoref{fig:lineplots-synthetic}e) and falling EOD for the Employment and Diabetes datasets (\autoref{fig:lineplots-synthetic}f, h).\footnote{Note that a higher value of EOD (\autoref{eq:eod}) signifies that females received more positive predictions than males.} The unconstrained classifiers (LR and RF) moved in the same direction for every dataset, either rising (\autoref{fig:lineplots-synthetic}e, f) or falling (\autoref{fig:lineplots-synthetic}h) with noise. The SOFT classifier also exhibited some variable behavior: on the Employment dataset EOD rose with noise (\autoref{fig:lineplots-synthetic}f), and on the Public Coverage (\autoref{fig:lineplots-synthetic}e) dataset it failed to achieve equal odds at higher noise levels. The remaining classifiers tended to achieve equal odds irrespective of the noise level.

\autoref{fig:lineplots-synthetic} only depicts average values for accuracy and EOD, which is potentially problematic because it may hide instability in the classifiers' performance. To address this we present \autoref{fig:stab-noise} in the Supplementary Material, which shows the distribution of accuracy and EOD results for each classifier on each dataset at the 0.1, 0.5, and 0.9 noise levels. We observe that, overall, no classifier became consistently less stable as noise increased. Rather, the stability patterns for each classifier mirrored the patterns that we already observed in \autoref{fig:baseline_stability}.

In summary, the classifiers that had problematic performance in the baseline experiments (see \autoref{fig:baseline_stability}) continued to have issues in the presence of noise. Additionally, the unconstrained classifiers exhibited inconsistent fairness as noise varied. Surprisingly, the noise-tolerant classifiers did not uniformly outperform the other fair classifiers.

\begin{figure*}[t]
    \centering
    \begin{subfigure}[t]{0.245\textwidth}
        \vskip 0pt
        \includegraphics[width=\columnwidth,keepaspectratio]{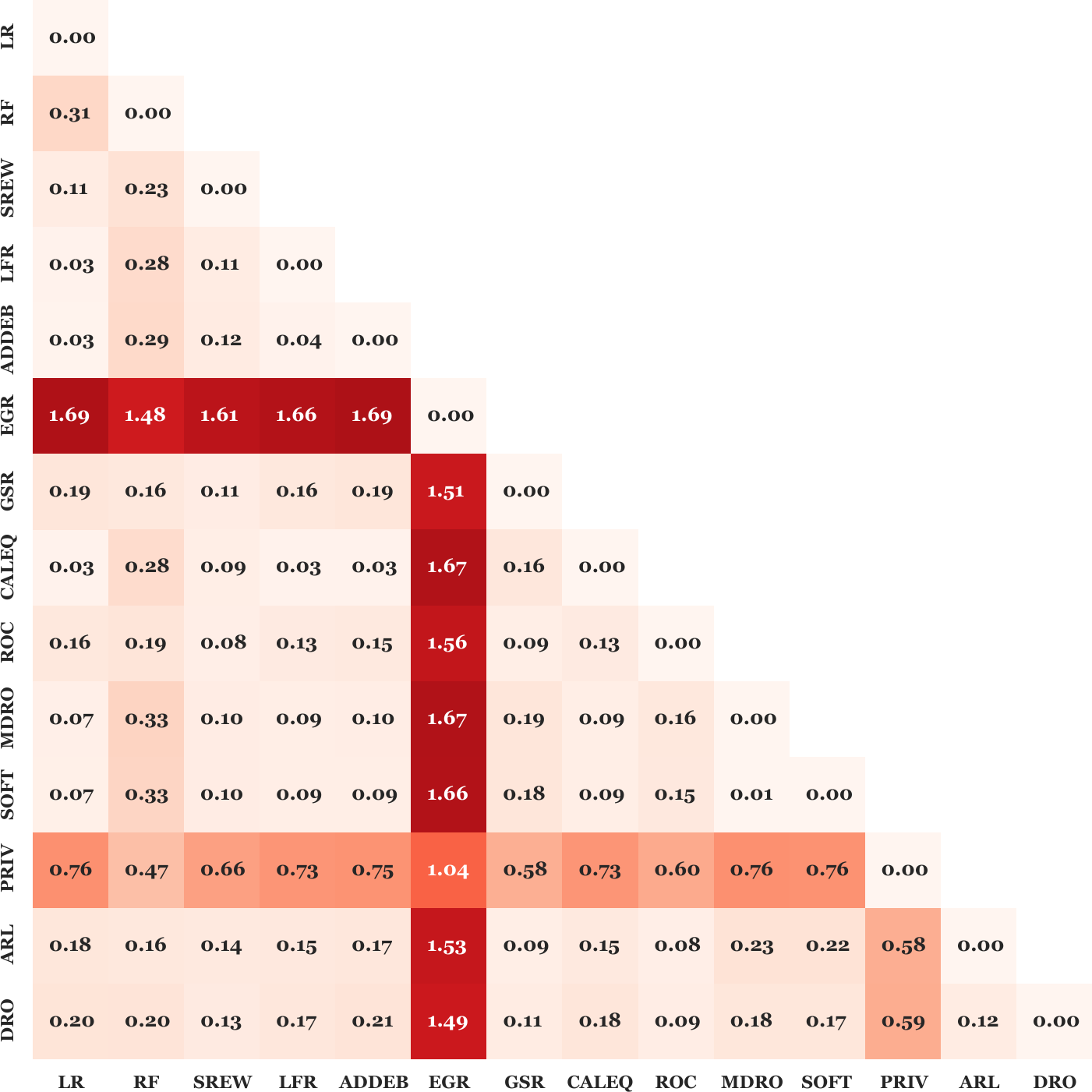}
        \caption{Public Coverage}
        \label{fig:distance-pub}
    \end{subfigure}
    % \hspace{10pt}
    \begin{subfigure}[t]{0.245\textwidth}
        \vskip 0pt
        \includegraphics[width=\columnwidth,keepaspectratio]{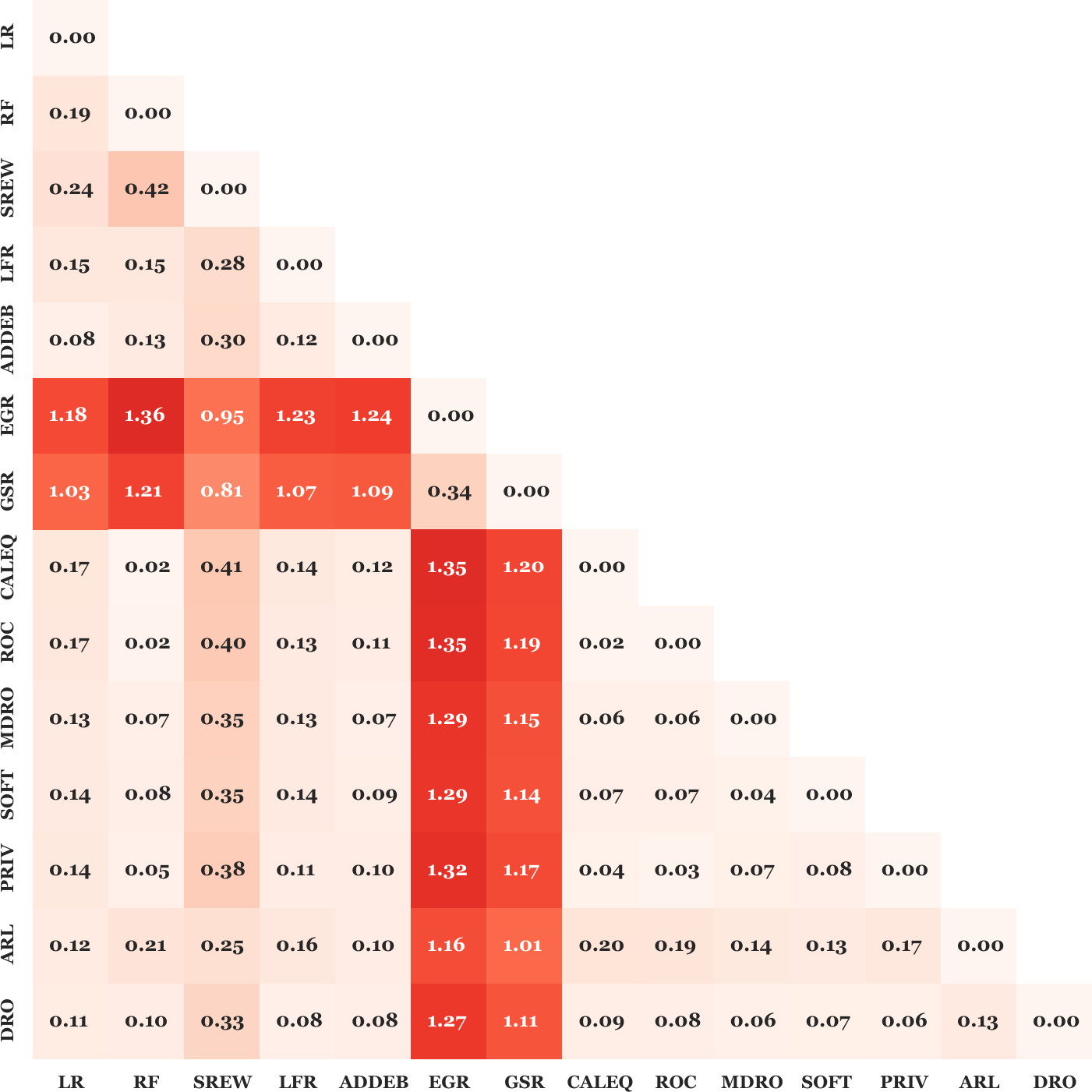}
        \caption{Employment}
        \label{fig:distance-emp}
    \end{subfigure}
    % \vfill
    \begin{subfigure}[t]{0.245\textwidth}
        \vskip 0pt
        \includegraphics[width=\columnwidth,keepaspectratio]{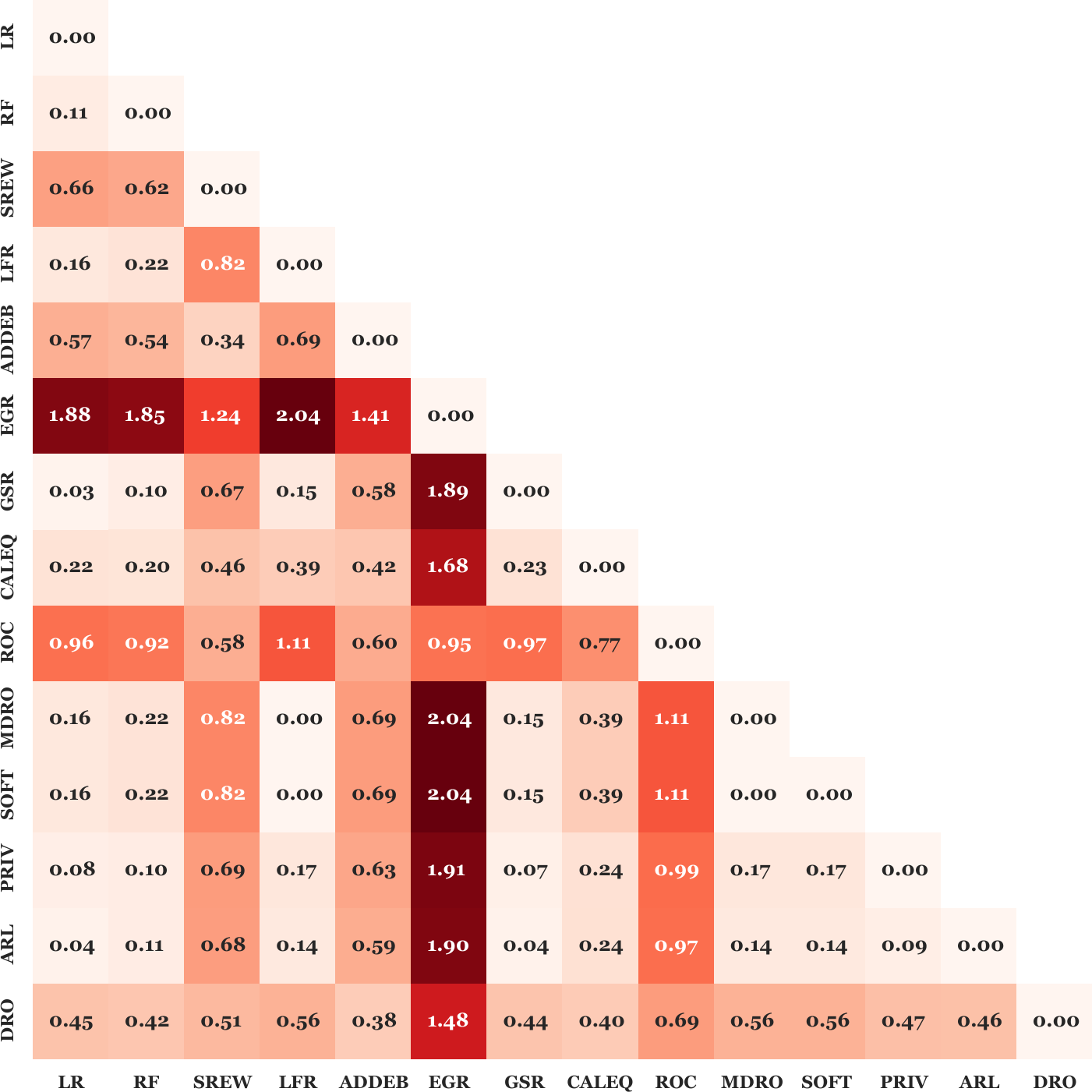}
        \caption{Law School Admissions}
        \label{fig:distance-law}
    \end{subfigure}
    % \hspace{10pt}
    \begin{subfigure}[t]{0.245\textwidth}
        \vskip 0pt
        \includegraphics[width=\columnwidth,keepaspectratio]{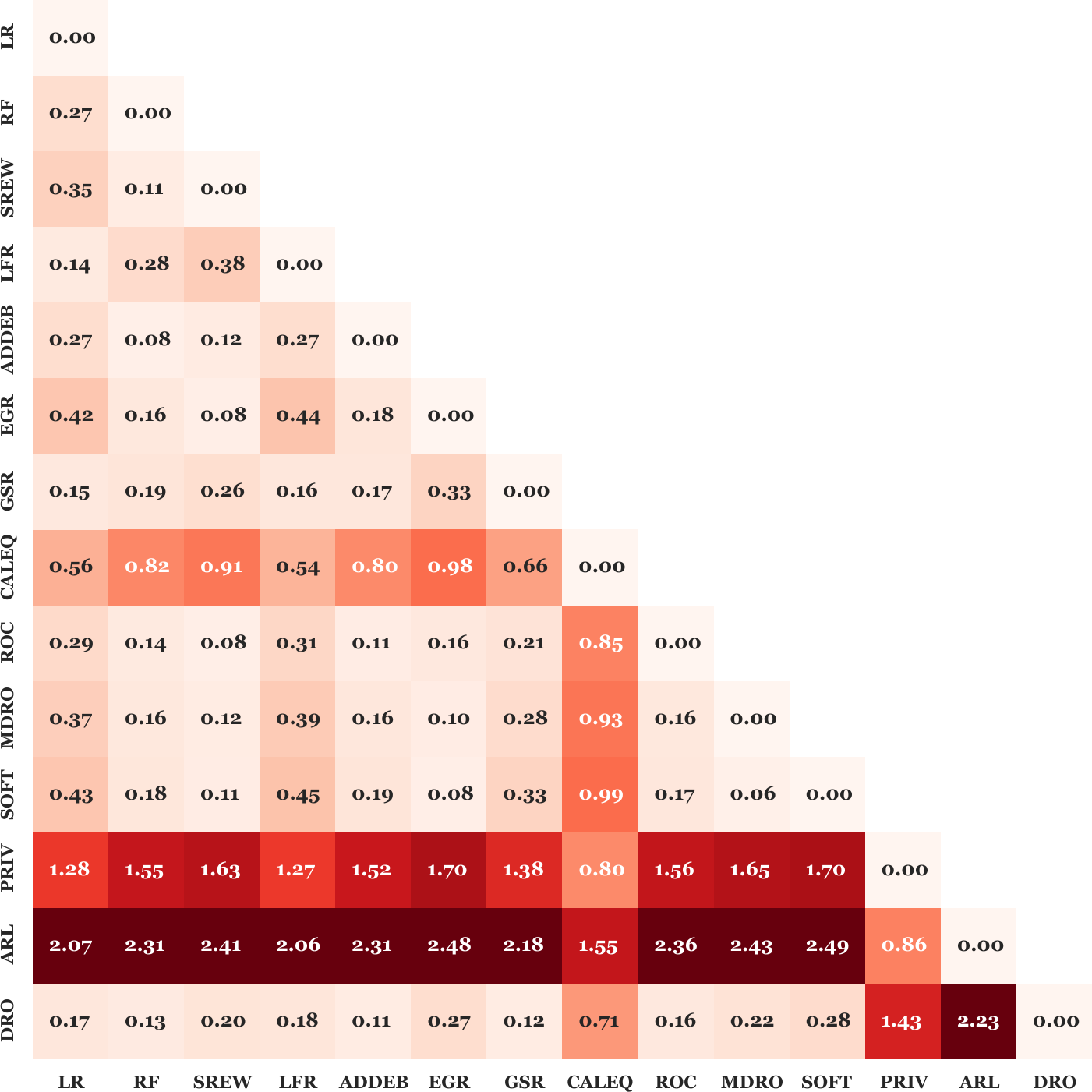}
        \caption{Diabetes}
        \label{fig:distance-dia}
    \end{subfigure}
    \vfill
    \caption{Wasserstein distances between the average KernelShap feature importance distributions over different noise levels for the four datasets. Each square compares the average feature importances of two classifiers. Redder squares denote pairs of classifiers with more divergent feature importance distributions.}
    \label{fig:distance}
\end{figure*}

\subsection{Feature Importance}

\begin{figure}[t]
    \includegraphics[width=\columnwidth,keepaspectratio]{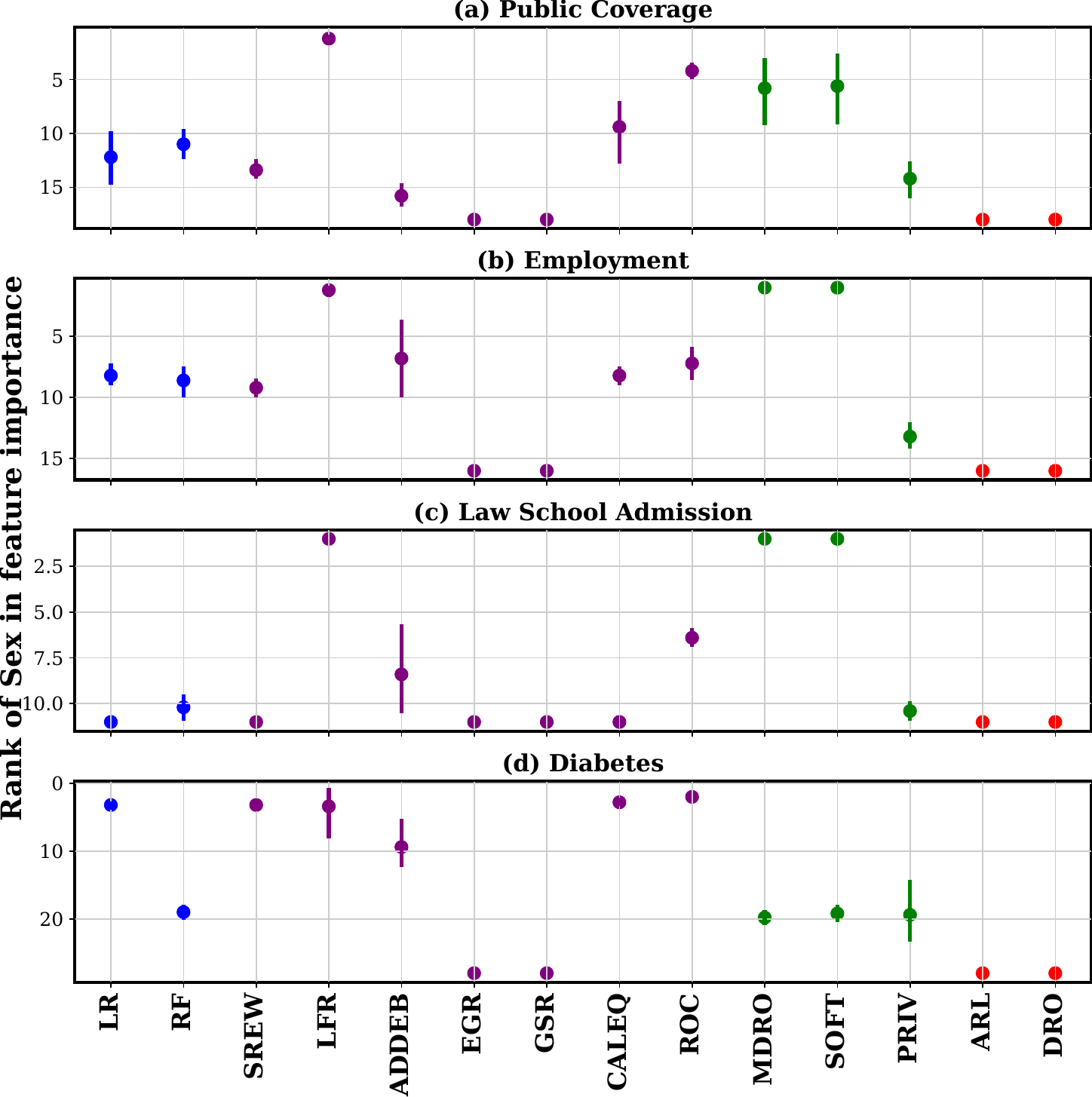}
    \caption{Rank of Sex in the average absolute KernelShap feature importances for the different algorithms in our case studies.}
    \label{fig:sex-rank}
    \vspace{-4mm}
\end{figure}

Finally, we delve into model explanations as a means to further explore the root causes of the classifier performance characteristics that we observed in the previous sections. First, we calculated feature explanations using KernelShap for every classifier at five noise levels---0, 0.2, 0.4, 0.6 and 0.8---using the method we described in \autoref{sec:meth:exp}. Next, we averaged the explanation distributions for each classifier to form a feature importance vector per classifier. Finally, we repeated this process for each dataset. For each dataset, we calculated Wasserstein distances~\cite{villani2009wasserstein} between the feature explanation distributions for each algorithm pair and present the results in \autoref{fig:distance}. Additionally, we plot the rank of the sex feature in terms of mean absolute feature importance for each classifier and present the results in \autoref{fig:sex-rank} (we also show the range of ranks if they vary over noise). 

\autoref{fig:distance} reveals that, with few exceptions (EGR in Public Coverage, EGR and GSR in Employment, EGR and ROC in Law school, and CALEQ, PRIV and ARL in Diabetes), most classifiers had similar feature explanation distributions. We do not observe any clear patterns among the exceptional classifiers, \ie no classifier consistently diverged from the others across all datasets. Further, we do not observe clear correlations between accuracy, EOD, and feature distribution similarity, suggesting that different classifiers took different paths to reach the same levels of performance.

\autoref{fig:sex-rank} is more informative than \autoref{fig:distance}. Four of the classifiers that exhibited consistently poor performance---LFR, MDRO, and SOFT (\autoref{fig:baseline_stability}a--d), and ROC (\autoref{fig:baseline_stability}e--h)---learned to weight the sex feature higher than other features, which may point to the root cause of their accuracy and fairness issues. Similarly, the unconstrained classifiers (LR and RF) exhibited changing EOD with noise levels in three out of four datasets (\autoref{fig:lineplots-synthetic}e, f, h), but not for Law School Admissions (\autoref{fig:lineplots-synthetic}g), and we observe that they learned a relatively low weight for sex among the available features for the Law School dataset. CALEQ also learned a relatively low weight for sex on the Law School dataset and was subsequently unaffected by noise (\autoref{fig:lineplots-synthetic}g), but showed variable trends in EOD for the other three datasets (\autoref{fig:lineplots-synthetic}e, f, h) on which it learned a relatively higher weight for sex.

Sex was the lowest ranked feature for the two demographic-unaware fair classifiers (DRO and ARL), which makes sense because they were not given these features as input. EGR and GSR also did not have access to sex while classifying the test dataset, so they also had sex as the lowest ranked feature.

\subsection{Fairness-Accuracy Tradeoff}

Three algorithms in our list - EGR, GSR, and PRIV, provide a mechanism to control the fairness-accuracy tradeoff via a hyperparameter -- namely fairness violation $eps$ in the case of EGR and GSR \cite{agarwal2018reductions}, and the privacy level $\epsilon$ in the case of PRIV \cite{mozannar2020fair}. Based on the experiments the authors of these algorithms did in their papers, we used different $eps$ values between 0.01 and 0.20 and $\epsilon$ values between -2 and 2 and reran our experiments. We found that tweaking the tradeoff hyperparameter did not contribute meaningfully to the stability and noise resistance capabilities of these algorithms. Consequently we omit these results from the paper.

\section{Conclusion}
\label{sec:conclusion}

In this study, we present benchmark results---in terms of accuracy, fairness, and stability---for 14 ML classifiers divided into four classes. We evaluated these classifiers across four datasets and varying levels of random noise in the protected attribute. Overall, we found that two classical fair classifiers (SREW and EGR), one noise-tolerant fair classifier (PRIV), and one demographic-unaware fair classifier (ARL) performed consistently well across metrics on our experiments. In the future we recommend that ML researchers benchmark their own fair classifiers against these classifiers and that practitioners consider adopting them.

One surprising finding of our study was how well SREW and EGR performed in the face of noise in the protected attribute. Contrast this to noise-tolerant classifiers like MDRO---whose performance did not vary with noise but was inaccurate on some datasets---and SOFT---which was consistently inaccurate and had variable fairness in the face of noise. These results suggest that some classical fair classifiers may actually fare well in the face of noise, and that adopting more complex noise-tolerant fair classifiers may not always be necessary.

Another surprising finding of our study was how well ARL performed. As a demographic-unaware fair classifier it did not have access to the sex feature at training or testing time, yet it achieved fairness performance that was comparable to demographic-aware fair classifiers on three of our datasets, and its fairness performance was noise invariant on three datasets as well. We fit linear regression models on each dataset with sex as the independent variable, but these models did not uncover any obvious proxy features for ARL to use in place of the sex feature. This speaks to the strength of the ARL algorithm's adversarial approach to learning.

On one hand, our results confirm that demographic-unaware fair classifiers can achieve fairness for real-world disadvantaged groups under ecological conditions. This is positive news for practitioners who would like to adopt a fair classifier but lack (high-quality) demographic data. On the other hand, we still urge caution with respect to the adoption of demographic-unaware fair classifiers for practical reasons. First, determining whether a classifier like ARL will achieve acceptable performance in a given context requires thorough evaluation on a dataset that includes demographic data, as we have done here. Second, even if a demographic-unaware fair classifier performs well in testing, its performance may degrade after deployment if the context changes or there is distribution drift~\cite{ghosh2022faircanary}. Monitoring the health of a classifier like ARL in the field requires demographic data. In short, adopting a demographic-unaware classifier does not completely obviate the need for at least some high-quality demographic data.

In general, the results of our study point to the need for further development in the areas of noise-tolerant and demographic-unaware fair classifiers. By releasing our source code and data, we hope to provide a solid foundation for evaluating these novel classifiers in the future.

Our study has several limitations. First, we only evaluate classifiers using binary protected attributes. It is unclear how their performance and consistency would change under more complex conditions. That said, we are confident that the classifiers that performed poorly will continue to do so in the presence of more complex fairness objectives. Second, our case studies and synthetic experiments, while thorough, are by no means completely representative of all real world datasets and contexts. We caution that our results should not be generalized indefinitely. Third, we did not evaluate all of the classical fair classifiers from the literature (see \citet{friedler2019comparative} and \citet{mehrabi2019survey} for more). Our primary focus was on adding to the literature by benchmarking noise-tolerant and demographic-unaware fair classifiers. Finally, in this study we only evaluated one fairness metric---EOD---because it was the common denominator among all of the classifiers we selected. Future work could explore fairness performance further by choosing other fairness metrics along with subsets of amenable classifiers.

% \section*{Acknowledgments}
\begin{acks}

We thank the anonymous reviewers for their helpful comments. We also thank Jeffrey Gleason for notes on the manuscript. This research was supported in part by \grantsponsor{1}{NSF}{https://www.nsf.gov/} grant \grantnum{1}{IIS-1910064}. Any opinions, findings, and conclusions or recommendations expressed in this material are those of the authors and do not necessarily reflect the views of the NSF.

\end{acks}

%%
%% The next two lines define the bibliography style to be used, and
%% the bibliography file.

\balance
\bibliographystyle{ACM-Reference-Format}
\bibliography{refs}

\clearpage

%%
%% If your work has an appendix, this is the place to put it.
\appendix

\section{Supplementary Material}

\begin{figure*}[b]
    \begin{subfigure}[t]{\columnwidth}
        \centering
        \vskip 0pt
        \includegraphics[width=\textwidth,keepaspectratio]{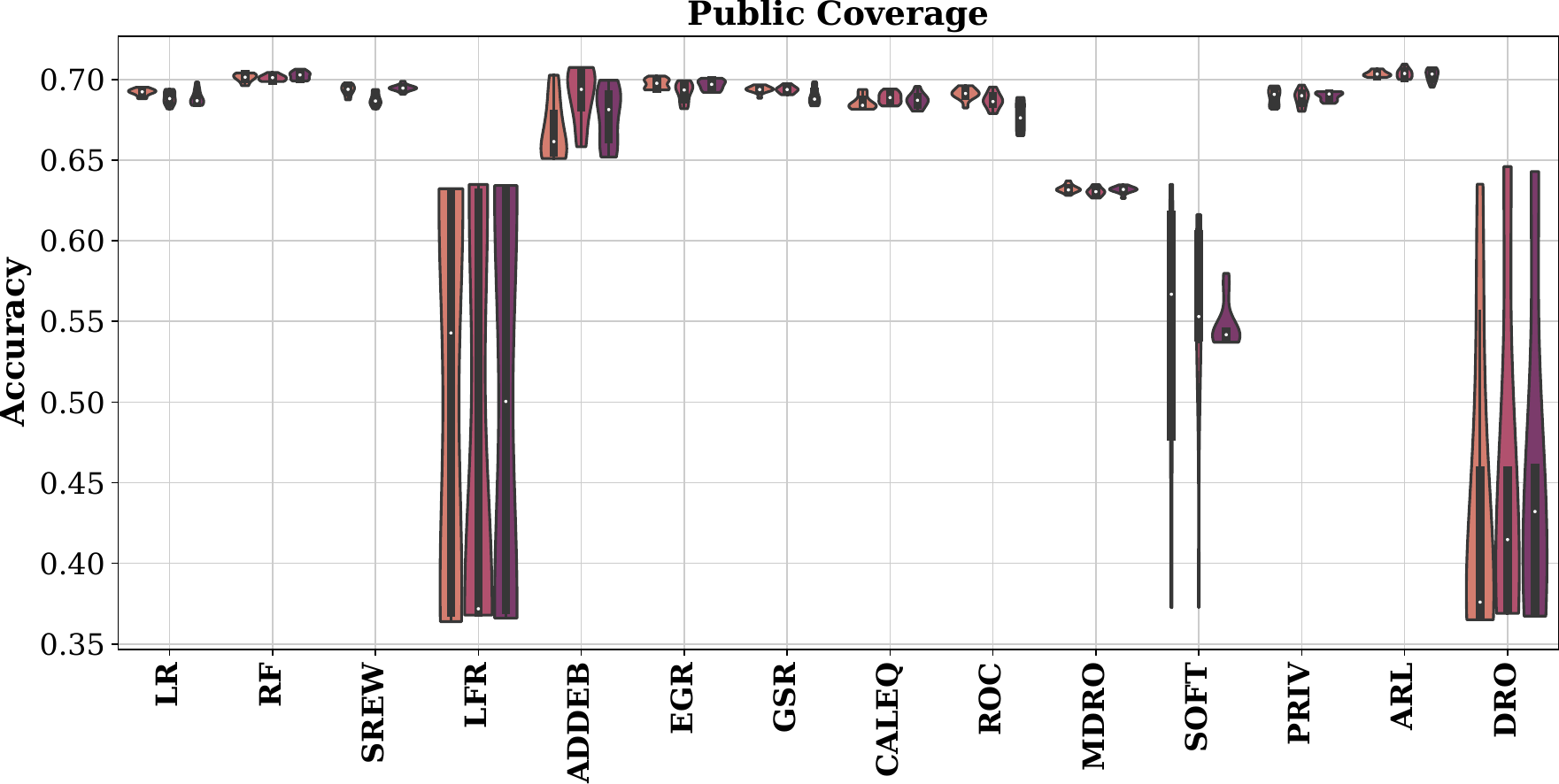}
    \end{subfigure}
    \hfill
    \begin{subfigure}[t]{1.03\columnwidth}
        \centering
        \vskip 0pt
        \hspace*{-13pt}
        \includegraphics[width=\textwidth,keepaspectratio]{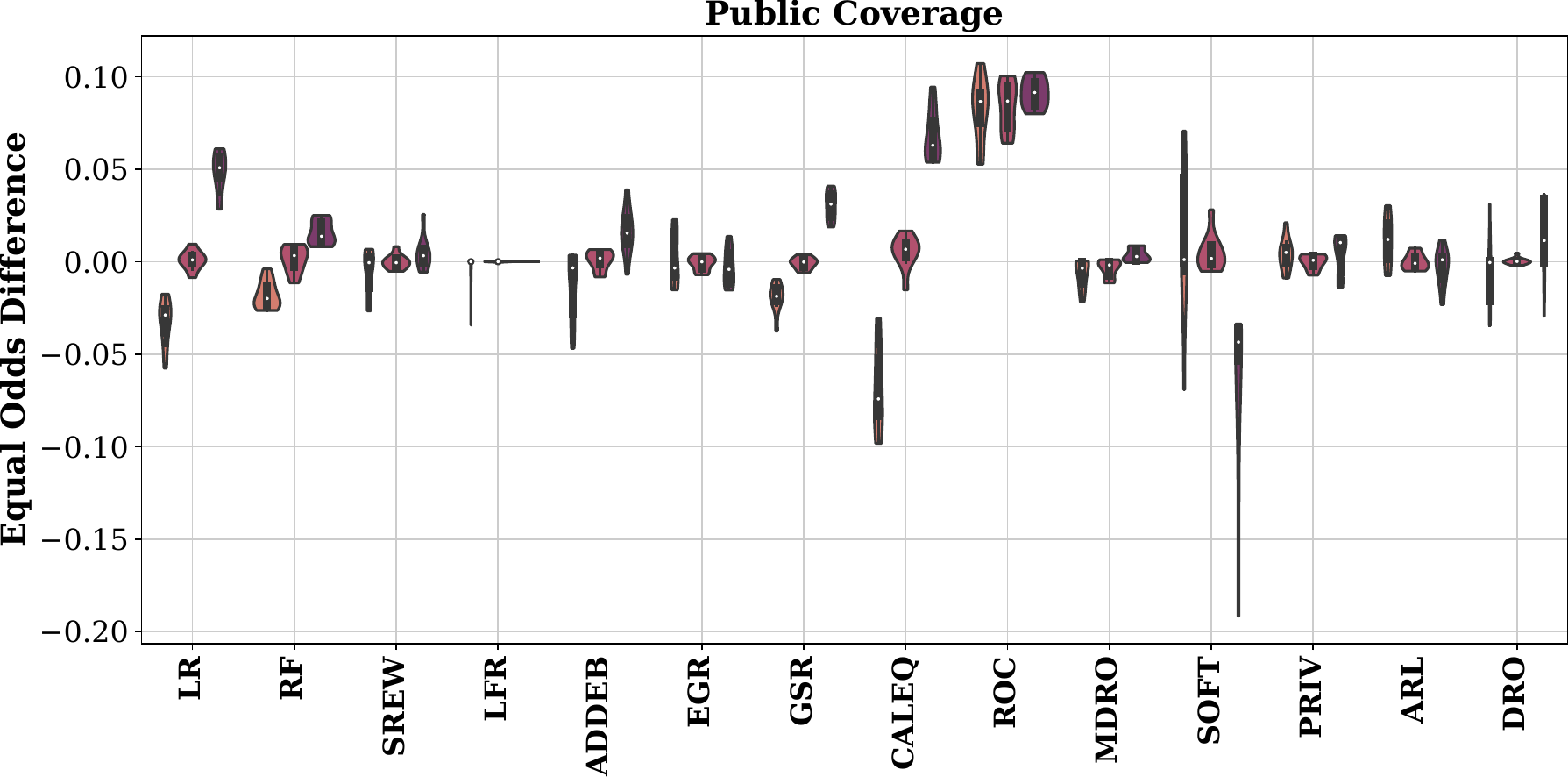}
    \end{subfigure}
    \vfill
    \begin{subfigure}[t]{\columnwidth}
        \centering
        \vskip 0pt
        \includegraphics[width=\textwidth,keepaspectratio]{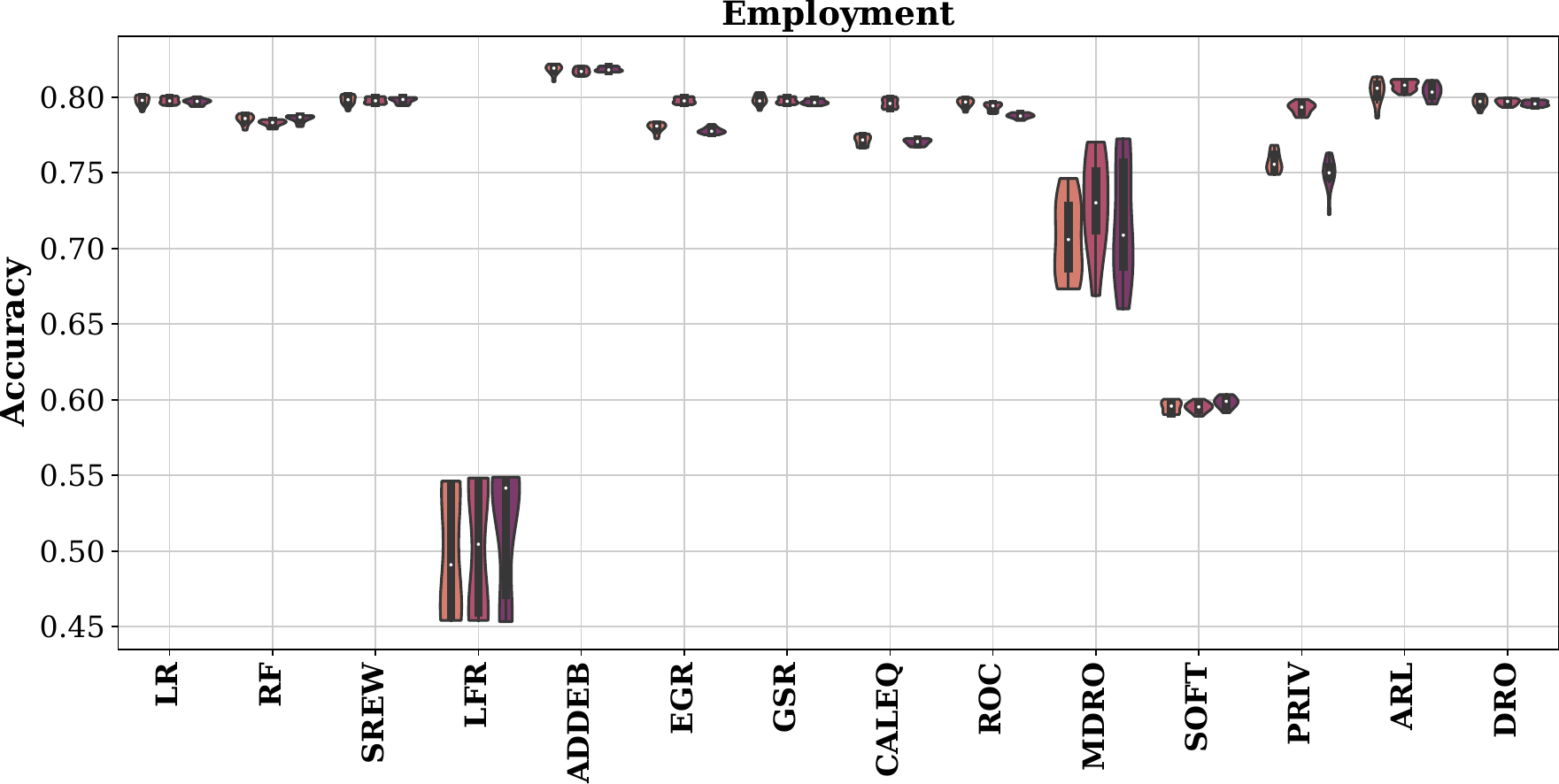}
    \end{subfigure}
    \hfill
    \begin{subfigure}[t]{1.03\columnwidth}
        \centering
        \vskip 0pt
        \hspace*{-13pt}
        \includegraphics[width=\textwidth,keepaspectratio]{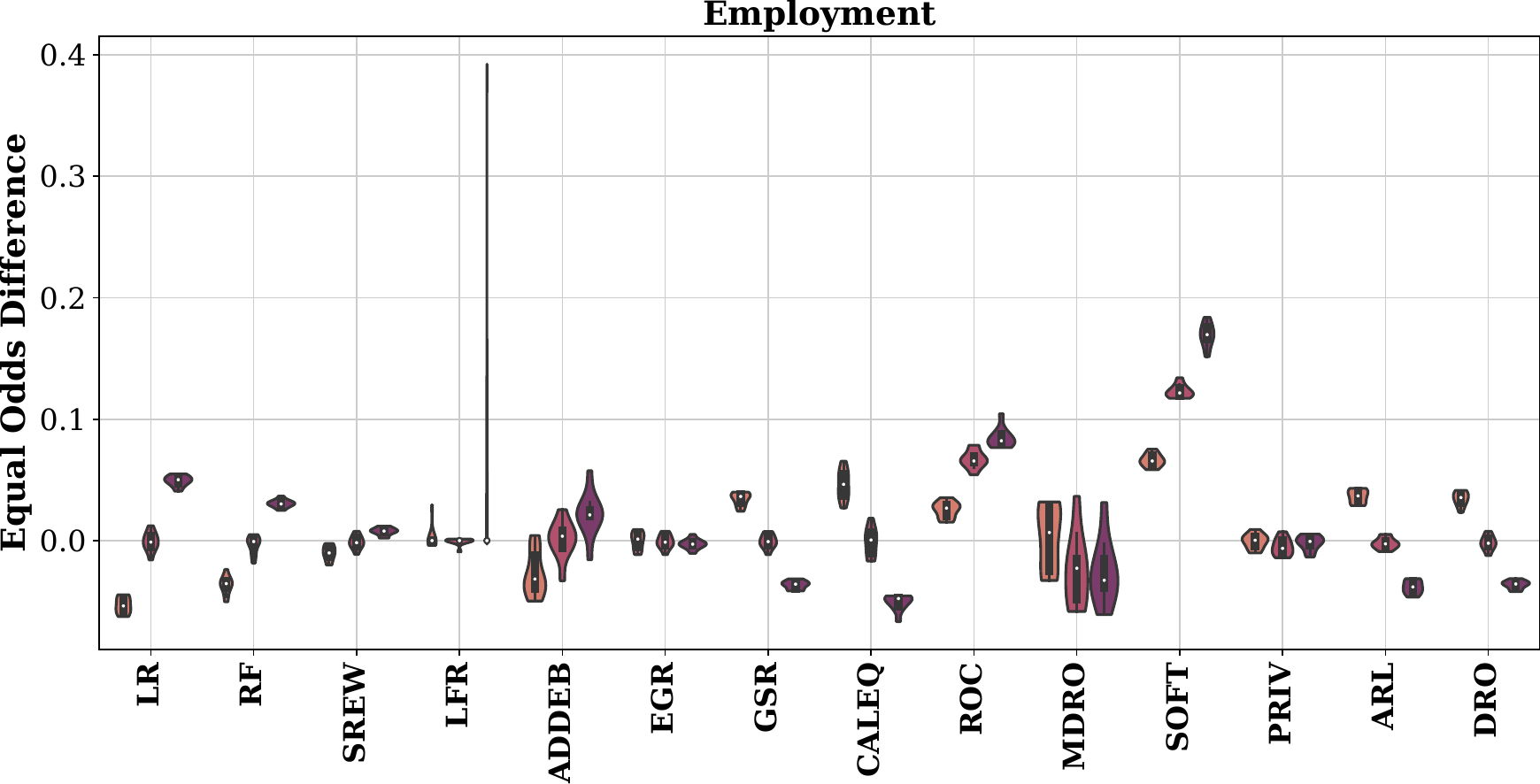}
    \end{subfigure}
    \vfill
    \begin{subfigure}[t]{\columnwidth}
        \centering
        \vskip 0pt
        \includegraphics[width=\textwidth,keepaspectratio]{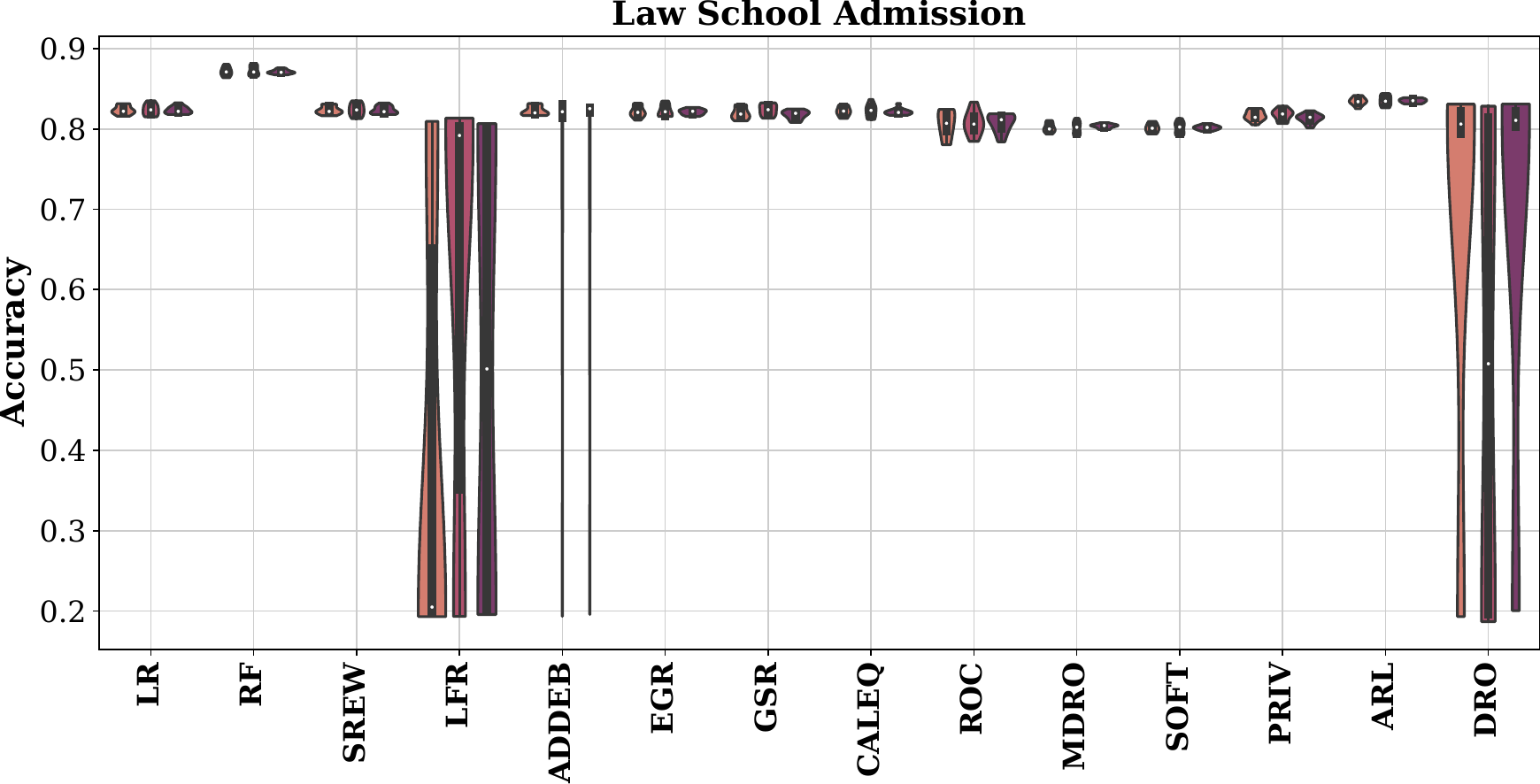}
    \end{subfigure}
    \hfill
    \begin{subfigure}[t]{1.03\columnwidth}
        \centering
        \vskip 0pt
        \hspace*{-13pt}
        \includegraphics[width=\textwidth,keepaspectratio]{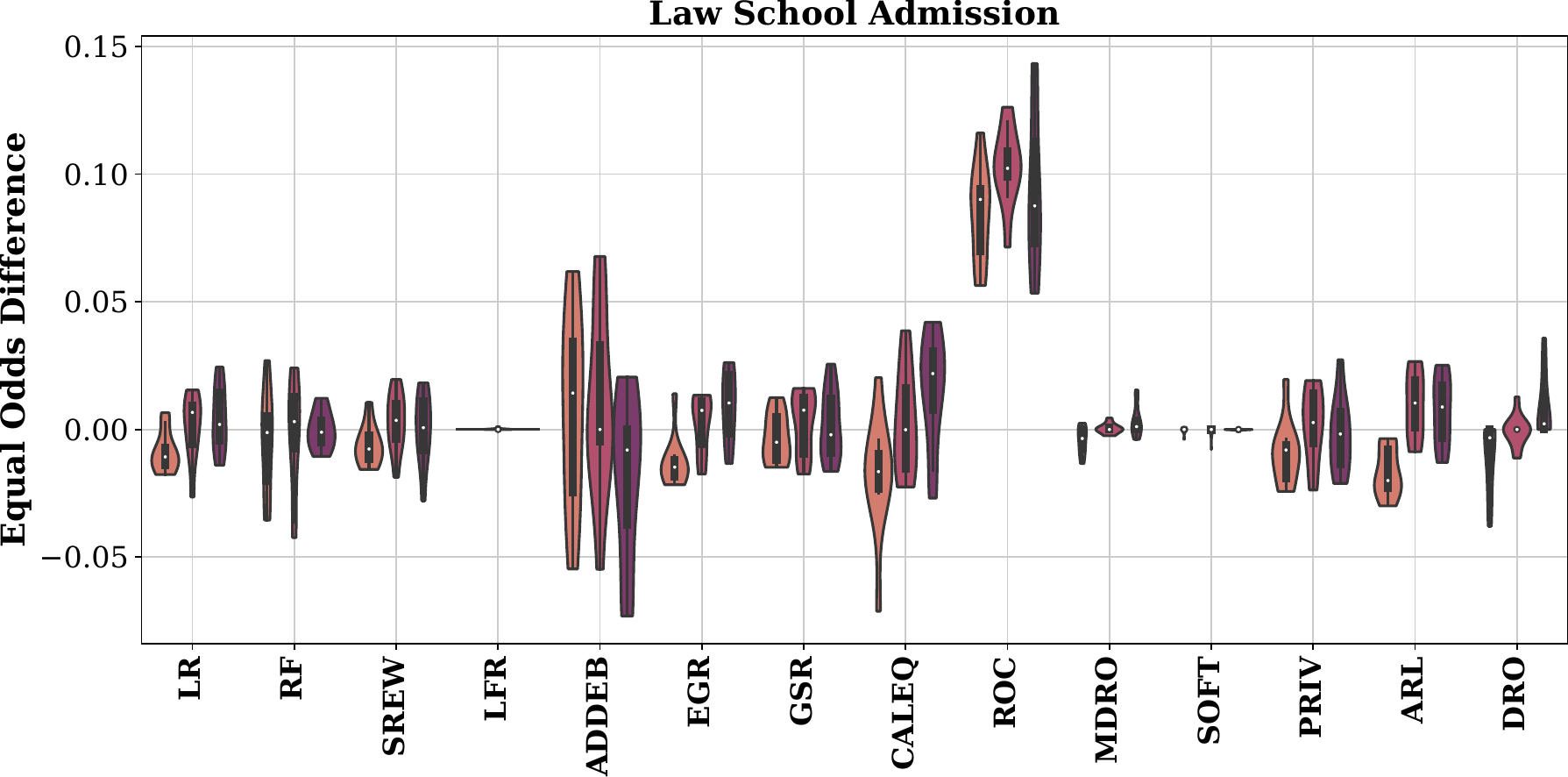}
    \end{subfigure}
    \vfill
    \begin{subfigure}[t]{\columnwidth}
        \centering
        \vskip 0pt
        \includegraphics[width=\textwidth,keepaspectratio]{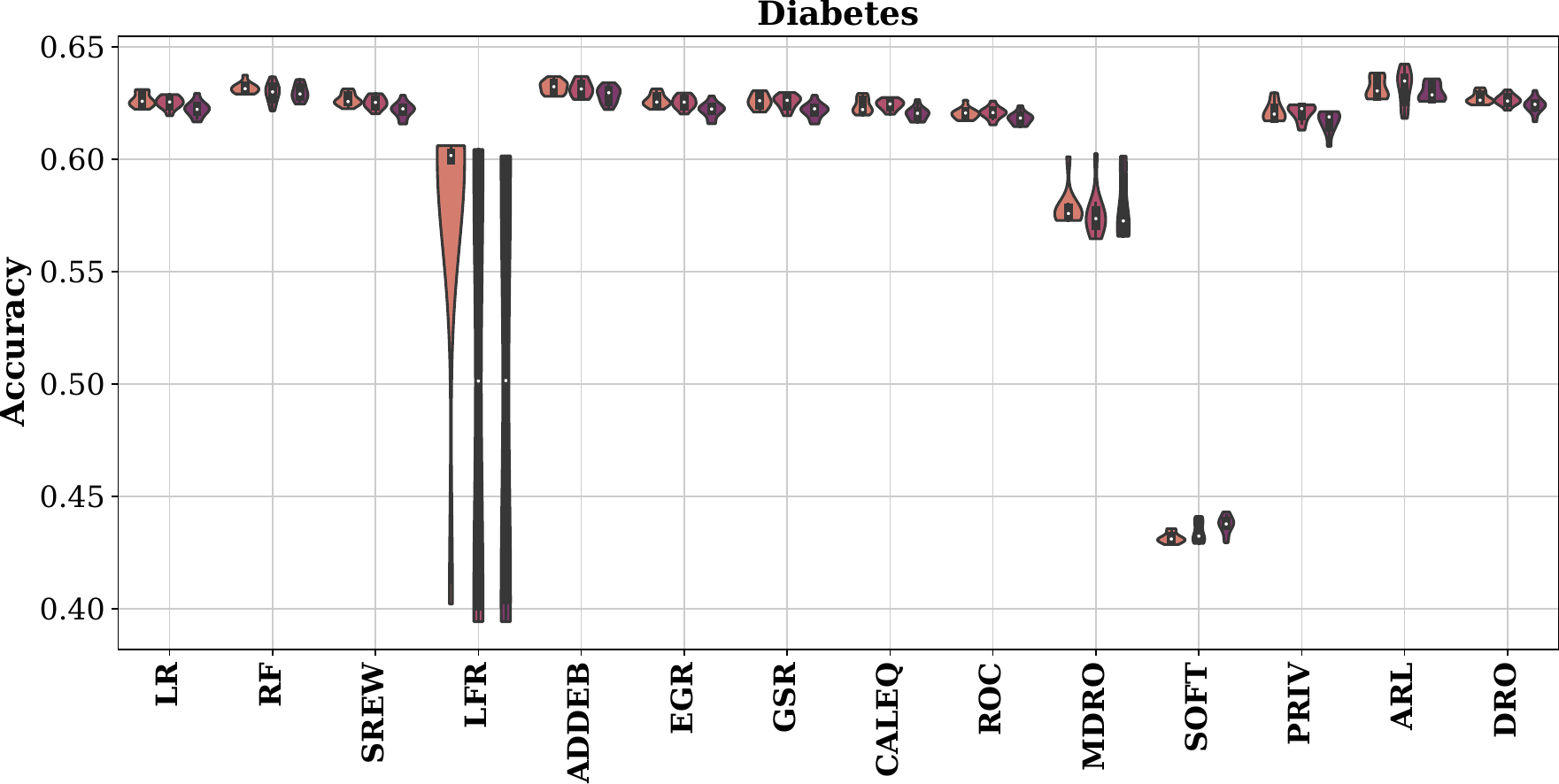}
    \end{subfigure}
    \hfill
    \begin{subfigure}[t]{1.03\columnwidth}
        \centering
        \vskip 0pt
        \hspace*{-13pt}
        \includegraphics[width=\textwidth,keepaspectratio]{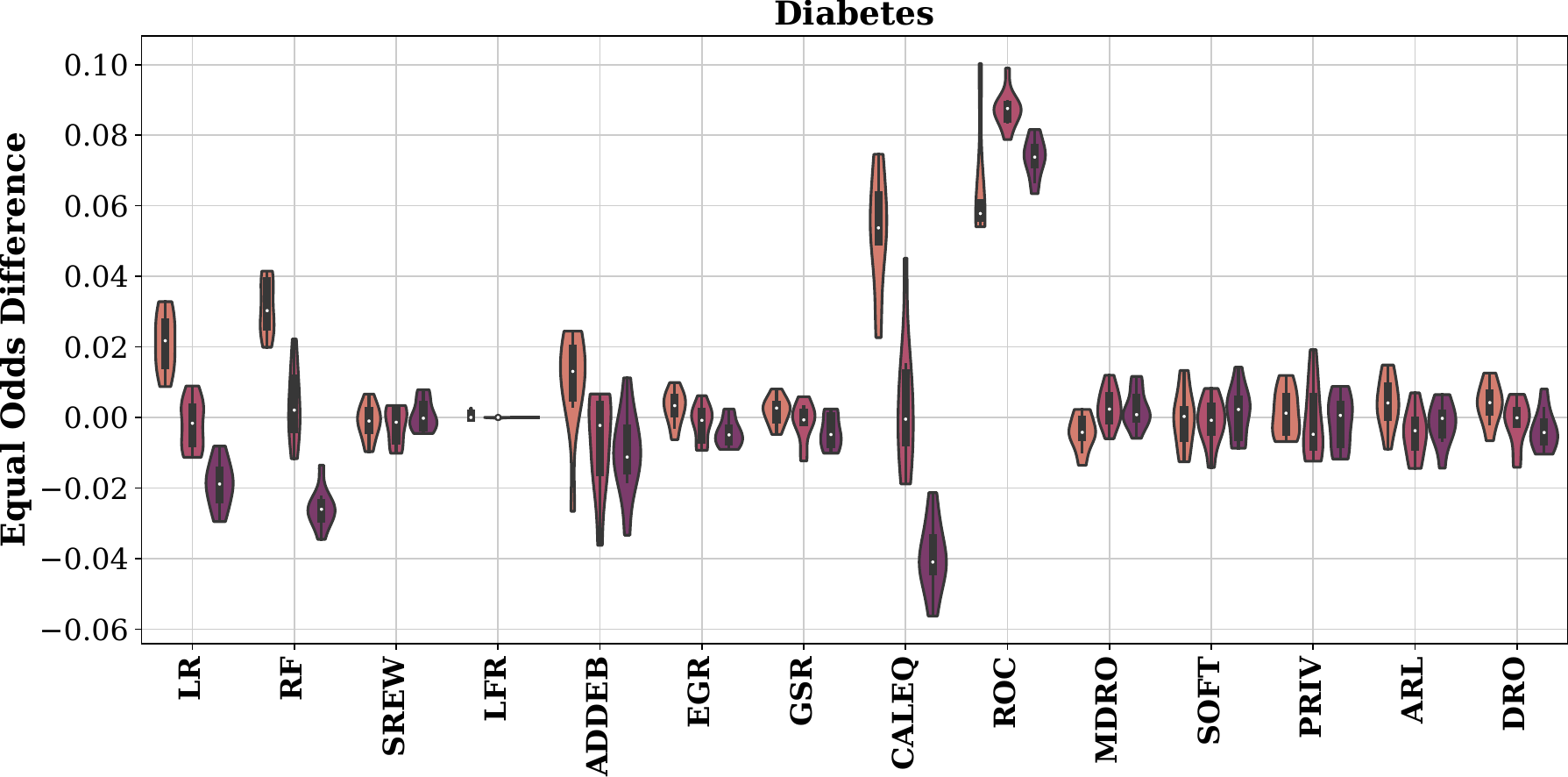}
    \end{subfigure}
    \vfill
    \caption{Plots showing the stability of our 14 classifiers over three different levels of noise in protected attributes (0.1, 0.5 and 0.9). For each dataset we present the stability of each classifiers' accuracy and EOD.}
    \label{fig:stab-noise}
\end{figure*}

% \subsection{Part One}

% Lorem ipsum dolor sit amet, consectetur adipiscing elit. Morbi
% malesuada, quam in pulvinar varius, metus nunc fermentum urna, id
% sollicitudin purus odio sit amet enim. Aliquam ullamcorper eu ipsum
% vel mollis. Curabitur quis dictum nisl. Phasellus vel semper risus, et
% lacinia dolor. Integer ultricies commodo sem nec semper.

% \subsection{Part Two}

% Etiam commodo feugiat nisl pulvinar pellentesque. Etiam auctor sodales
% ligula, non varius nibh pulvinar semper. Suspendisse nec lectus non
% ipsum convallis congue hendrerit vitae sapien. Donec at laoreet
% eros. Vivamus non purus placerat, scelerisque diam eu, cursus
% ante. Etiam aliquam tortor auctor efficitur mattis.

% \section{Online Resources}

% Nam id fermentum dui. Suspendisse sagittis tortor a nulla mollis, in
% pulvinar ex pretium. Sed interdum orci quis metus euismod, et sagittis
% enim maximus. Vestibulum gravida massa ut felis suscipit
% congue. Quisque mattis elit a risus ultrices commodo venenatis eget
% dui. Etiam sagittis eleifend elementum.

% Nam interdum magna at lectus dignissim, ac dignissim lorem
% rhoncus. Maecenas eu arcu ac neque placerat aliquam. Nunc pulvinar
% massa et mattis lacinia.

\end{document}